\documentclass[12pt]{article}
\usepackage{arxiv}

\usepackage[utf8]{inputenc} % allow utf-8 input
\usepackage[T1]{fontenc}    % use 8-bit T1 fonts
\usepackage{hyperref}       % hyperlinks
\usepackage{url}            % simple URL typesetting
\usepackage{booktabs}       % professional-quality tables
\usepackage{amsfonts}       % blackboard math symbols
\usepackage{nicefrac}       % compact symbols for 1/2, etc.
\usepackage{microtype}      % microtypography
\usepackage{graphicx}
\usepackage{amsmath}
\usepackage{enumitem}
\usepackage{subcaption}

\title{Deep Visual Re-Identification with Confidence}
\author{
  George Adaimi\\%\thanks{Use footnote for providing further
  VITA, EPFL\\
  Switzerland\\
  \texttt{george.adaimi@epfl.ch} \\
   \And
 Sven Kreiss \\
  VITA, EPFL\\
  Switzerland\\
  \texttt{sven.kreiss@epfl.ch} \\
  \And
  Alexandre Alahi\\
  VITA, EPFL\\
  Switzerland\\
  \texttt{alexandre.alahi@epfl.ch} \\
}

\date{}
\begin{document}

\maketitle

\begin{abstract}
    Transportation systems often rely on understanding the flow of vehicles or pedestrian. From traffic monitoring at the city scale, to commuters in train terminals, recent progress in sensing technology make it possible to use cameras to better understand the demand, \textit{i.e.}, better track moving agents (\textit{e.g.}, vehicles and pedestrians). Whether the cameras are  mounted on drones, vehicles, or fixed in the built environments, they inevitably remain scatter. We need to develop the technology to re-identify the same agents across images captured from non-overlapping field-of-views, referred to as the visual re-identification task. State-of-the-art methods learn a neural network based representation trained with the cross-entropy loss function. We argue that such loss function is not suited for the visual re-identification task hence propose to model confidence in the representation learning framework. We show the impact of our confidence-based learning framework with three methods: label smoothing, confidence penalty, and deep variational information bottleneck. They all show a boost in performance validating our claim. Our contribution is generic to any agent of interest, \textit{i.e.}, vehicles or pedestrians, and outperform highly specialized state-of-the-art methods across 5 datasets. The source code and models are shared towards an open science mission.

\end{abstract}

% keywords can be removed

\begin{figure}[t]
\begin{center}
\includegraphics[width=\linewidth]{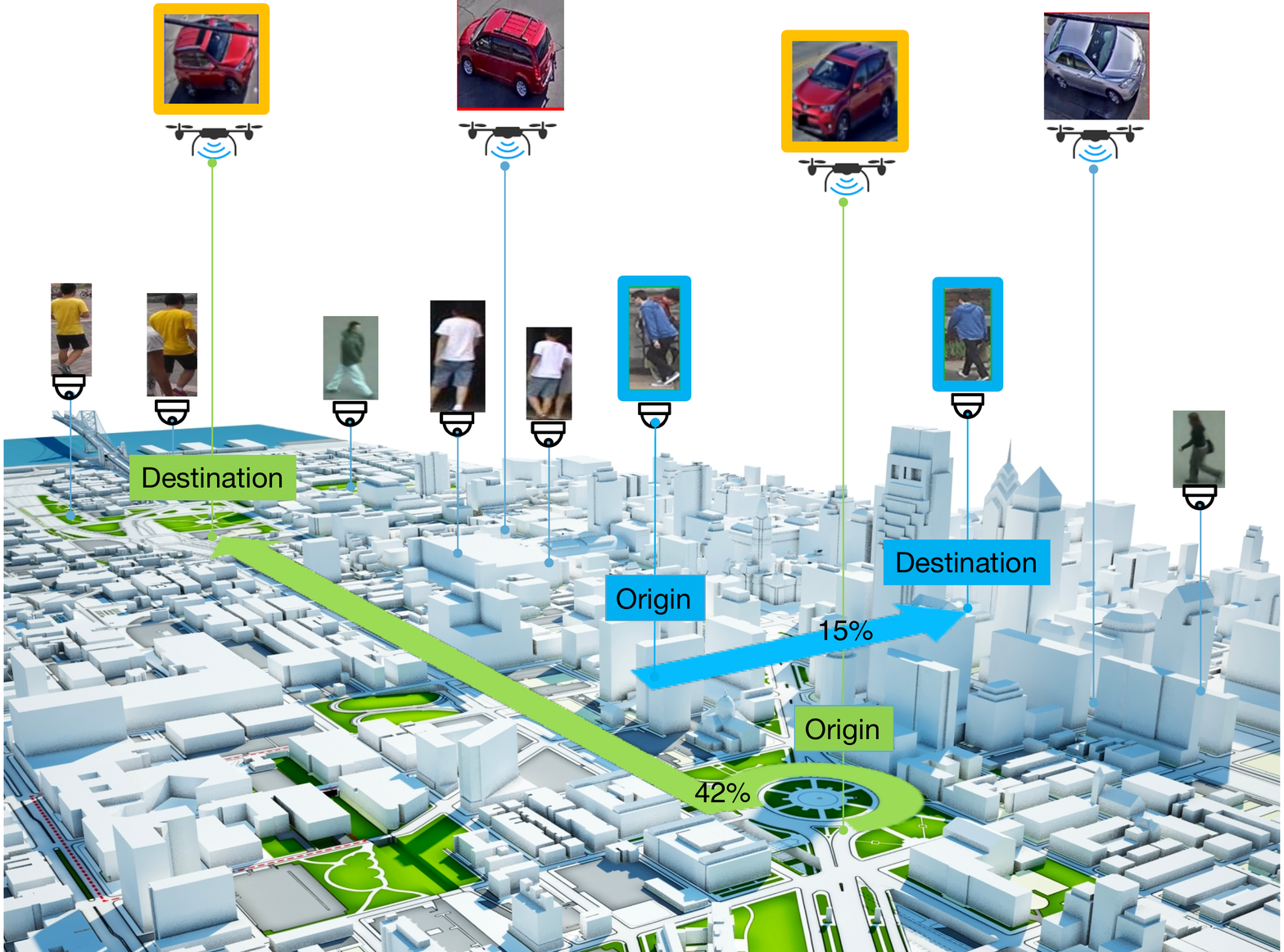}
\end{center}
\caption{In this work, we present a new visual re-identification method, \textit{i.e.}, whether agents (vehicles or pedestrians) captured in different images in non-overlapping views belong to the same agent. Agents with the orange bounding box belong to the same agent. Note different agents can be visually very similar.}
\label{fig:pull}
\end{figure}

\section{Introduction}
An important goal of transportation research is to improve and provide efficient public transportation systems that can accommodate many agents, whether it be vehicles or pedestrians, every day. This is especially important nowadays with the huge traffic congestion costing billions of dollars~\cite{schneider_2018}. As a result, research efforts have been directed towards management and control of vehicle and pedestrian flows. Important prerequisites for such transportation network analysis are origin-destination (OD) matrices, which allow researchers to understand a population's trip demand. With the recent developments in new methods, such as data-driven methods~\cite{krishnakumari2019data}, OD estimation have achieved good performance in various tasks of traffic management. However, it still faces the issue of how representative the chosen samples are of the population. One way to deal with this problem is to collect more data from the population, which is expensive and time-consuming when using traditional methods such as surveys or interviews. Another way is to make use of smartphones to collect such data~\cite{zhao2015stop}. Visual re-identification is a faster and cheaper way to collect these data. Visual re-identification is the task of associating images of the same agent taken from different cameras or from the same camera in different occasions. This is represented in Fig.~\ref{fig:pull}. This task is nowadays possible due to the complex network of cameras already places in and around cities. Moreover, recent works have been pushing towards the use of drone technology to collect massive amounts of data in order to study the traffic phenomena, such as the recently released pNEUMA dataset~\cite{barmpounakis2020new}. All these collected data can be used as inputs to a visual re-identification model to associate different agents together and obtain their paths. %%Another benefit of this task is that it can be used to improve the navigation of self-driving cars and service robots. Over many years, people learned to easily navigate through our complex environment by subconsciously using different cues and changes. For autonomous vehicles to do the same, they need to keep track of the agents surrounding him over time to better predict future changes and to understand pedestrian's complex behaviors.

The task of re-identification (re-id) has long been a task of extracting features/representations from two observations and measuring how similar these features are. Since different variations affect these features, many works have introduced different methods to improve their extraction. Initially, these features were hand-crafted and include spatio-temporal information such as color, width and height, and salient edge histograms. Some work have also tried to use different input modalities such as depth~\cite{7864367,Karianakis2017PersonDR,7486459}, infrared~\cite{Wu_2017_ICCV}, LiDAR~\cite{TRC_LiDAR}, or Inductive Loop Detectors for vehicles~\cite{abdulhai2003spatio}. These features, however, fail drastically when dealing with unexpected scenarios. To remedy this problem and with the advent of machine learning, researchers are now benefiting from the strength of deep learning to be able to extract more general and more discriminative features allowing them to reach high performance. Since then, an arms race of methods was built on top of this by making use of different object-specific characteristics (\textit{e.g.}, human semantic segmentation, pose) and by learning features through the supervision of a cross-entropy loss.

A main pitfall of learning with cross-entropy supervision is the fact that it separates the different inputs solely based on the labels without taking into consideration the actual similarity between the inputs. None of the recent methods have tackled the problem where, even though two very similar agents are distinct, their similarity score should encode information about how similar they appear while also distinguishing them. The network usually tries its best to find a boundary between the different classes even for inputs that are very similar. This leads the network to find unreasonable explanations for the differences in labels and thus would negatively affect its generalizability. Since re-id also deals with the problem of having a small set of images per class, it would aggravate this issue. Controlling the network's confidence in its predictions would alleviate this problem. To the best of your knowledge, this is the first paper to apply this concept in the field of re-identification.

In this paper, we propose to model confidence when learning representations appropriate for re-id. By inducing doubt while training a network, we are able to tackle the inherent problem discussed previously when cross-entropy is used in a distance metric and representation learning problem such as person or vehicle re-identification. Inspired by previous works that use uncertainty to regularize the network, we study three alternatives that aim at reducing the confidence of the network and show a gain of 6-7 \% in mAP across 5 different datasets. Although these methods have shown only a small improvement in other image classification tasks~\cite{Szegedy_2015_CVPR,DBLP:journals/corr/AlemiFD016,DBLP:journals/corr/PereyraTCKH17}, they drastically improve the performance of re-id models due to its innate problem (Section~\ref{sssec:problem}). By combining our methods with advanced ranking methods, we outperform state-of-the-art models without modeling additional characteristics specific to the object in question.  The software is open-source and is made available online\footnote[1]{https://git.io/deep-visual-reID-confidence}.

\section{Related Work}

Initially, re-identification's origins were based on multi-camera tracking. In this context, both visual and spatio-temporal information were used to predict whether the same agent was found in different cameras and at different times~\cite{WANG20133,Huang:1997,mazzon2012person,held2014combining}.  Gheissari et al.~\cite{1640938} were the first to make use of only visual information to match different pedestrians. They made use of color and salient edge histograms to re-identify around 44 pedestrians recorded by 3 different cameras. This work clearly divided visual re-identification from multi-camera tracking. Alahi et al.~\cite{alahi2014robust} showed the benefits of visual re-identification in a network of fixed and mobile cameras. Fixed cameras installed at urban intersection can help the detection algorithms of cameras mounted on vehicles. Since then, researchers have proposed many methods to solve the visual re-identification tasks\cite{corvee2010person,corvee2010personhaar}. In the remaining of this section, we briefly present key methods tackling the problem for "person" and "vehicle".

\subsection{Person Re-Identification}

Even though person re-identification is highly beneficial to analyze how people make use of different transportation modes, it is not a widely studied matter in this context. Recent works in the transportation field have been using traditional and linear methods such as the Kalman filter~\cite{TRC_LiDAR, doi:10.5772/62758} and particle filters~\cite{8229782, particleFilter,hue2002tracking} to track pedestrians by using spatio-temporal information. These methods work well when tracking for a short period and while assuming that pedestrian's movement is linear. Applying the same methods over multiple cameras as well as in more complicated environments such as in a train station would lead to poor identification of passengers. %%%when people's movement is expected

With the prevalence of deep neural networks in most computer vision tasks, person re-id followed this success when Li et al.~\cite{Li_2014_CVPR} introduced a deep learning method for re-id that tried to overcome the problems of bounding box misalignment, photometric and geometric transforms while also introducing a new bigger dataset specifically for this task. This paved the way for new methods and datasets to emerge, causing the person re-id performance of machines to improve. Other work developed new methods that tackle specific challenges in person re-id by introducing different architectures and modules~\cite{Li_2017_CVPR,Li_2018_CVPR,Song_2018_CVPR,Wang_2018_CVPR}.%\vspace{-5pt}
\begin{description}[style=unboxed,leftmargin=0cm]

\item[Attention in Person reID.]%Attention
Recently, many methods have tried to improve the representation of the input by training multiple networks that extract global and local features and then combining these features to form the final representation.  This is usually done by using either a deterministic way of dividing the different parts of the representation \cite{Yi_2014_ICPR,Cheng_2016_CVPR,Varior2016,Ahmed_2015_CVPR} or making use of attention modules to separate the different parts~\cite{Li_2018_CVPR,Zhao_2017_ICCV,DBLP:journals/corr/YaoZZLT17}. Other works extracted intermediate representations to gain information about the input at different levels arguing that this allows the network to learn distinctive characteristics of the input at different scales~\cite{Chang_2018_CVPR,Wang_2018_ECCV, Wang_2018_CVPR}. Even though these methods showed improvement over their predecessors, they usually require separate networks to process each of the different features, leading to a more complex architecture and training procedure.
\item[Human Characteristics.]% Human parsing such as pose, keypoints
Another direction other researchers have taken is to make use of information and characteristics related specifically to humans in order to improve person re-id. The work by Xu et al.~\cite{Xu_2018_CVPR} aims at detecting three different types of pose information such as keypoints, rigid body parts (e.g., torso), and non-rigid parts (e.g., limbs). These information were extracted using an off-the-shelf human pose estimator. Then, with the help of these body parts, the features extracted by a feature extractor are refined and used to classify the different pedestrians. The use of third-party methods makes their model highly dependent on the performance of these methods. Another approach by Sarfraz et al~\cite{Sarfraz_2018_CVPR} uses keypoint information, in addition to the input image, to train a ResNet-50 model as well as another connected module that detects the view (front, back or side). Kayaleh et al.~\cite{Kalayeh_2018_CVPR} also made use of features extracted from different body parts and concatenated them to form a global feature which in turn was used to perform re-identification. The disadvantage of these methods is their high dependence on other methods and datasets that require annotation. Moreover, the fact that these models depend on specific human characteristics prevents them from being leveraged for other image-retrieval and clustering tasks.
\item[Re-Ranking.]% G2G and P2G + Re-Ranking
In addition to learning better features, many works have tried to improve the ranking process of person re-id by including information about how the different galleries are related instead of just using the relationship between the pairs of queries and galleries~\cite{Zhong_2017_CVPR,Ye:2015:ROP:2733373.2806326,Shen_2018_CVPR,Zhong_2017_CVPR,7410511, Leng:2015:PRC:2820479.2821725,Ye:2015:ROP:2733373.2806326,7557057}. Zhong et al~\cite{Zhong_2017_CVPR} introduced a method for refining the distances between the queries and galleries by making use of the k-reciprocal nearest neighbors. This is done as a post-processing step to improve the ranking process. Shen et al.~\cite{Shen_2018_CVPR} argued that this does not help in learning better features during training and introduced a new learnable module that performs a random walk on a graph connecting the different gallery images. By performing a random walk operation, gallery-to-gallery (G2G) information is taken into consideration while training the network, thus resulting in a more complete representation that provides a better ranking performance. Other methods also tried to include G2G information by using Graph Neural Networks~\cite{Shen_2018_ECCV} and Conditional Random Fields~\cite{Chen_2018_CVPR}. We will make use of G2G information by applying different re-ranking methods.
\item[Metric Learning.]
Several previous works have tried to tackle the problem of person re-id by introducing new metric loss functions. Both contrastive~\cite{1640964} and binary loss functions have been employed in order to push apart negative image pairs while pulling positive image pairs together~\cite{Varor_2016_eccv,Ahmed_2015_CVPR}. Taking into consideration both the pull and push of contrastive loss, other methods\cite{Wang_2018_CVPR, Liu_2018_CVPR,Wang_2018_ECCV} used triplet loss that simultaneously tackles negative and positive pairs leading to a less greedy method. Chen et al.~\cite{Chen_2017_CVPR} extended this loss to quadruplet inputs. The drawback of these methods is their high sensitivity to the sampling technique used. As a result, Yu et al.\cite{Yu_2018_ECCV} introduced the HAP2S loss to tackle this drawback and showed improvement in performance. All the above methods try to encode metric information in the embedding space compared to cross-entropy which is considered as a representation learning method.
\end{description}
\subsection{Vehicle Re-Identification}

Vision-based methods for vehicle re-identification are rather new. Initially, vehicle datasets were small and mainly used for car color, model classification, or detection~\cite{tian2015vehicle}. As a result, Liu et al.~\cite{liu2016large} built a large scale dataset for vehicle re-identification similar to person re-id datasets. They also made use of many techniques derived for person re-id and compared their performance. They later extended their work and added license plate verification as a way to improve re-identification~\cite{liu2016deep}. This intuition comes from the fact that each car has a unique license plate. Metric learning techniques were also extended for this task. Liu et al~\cite{liu2016deeprelative} introduced coupled clusters loss, a variant of triplet loss, to deal with sensitivity to the choice of the triplet samples. Zhang et al.~\cite{zhang2017improving} also improved triplet loss by augmenting the training with a classification loss and modifying the dataset sampling method. Instead of randomly sampling triplets of anchors, positive, and negative samples, their method ensure that the negative sample is an anchor or positive sample in another triplet. This provides a way for negatives to be pushed towards similar images rather than being pushed away from the anchor randomly. Bai et al.~\cite{bai2018group} tried to deal with the problem of inter-class similarity and intra-class variance by introducing the group sensitive triplet embedding (GSTE). This is done by combining samples into intermediate "groups" at different granularity levels such as vehicle ID and vehicle model.

Other works made use of different attributes and modalities to improve vehicle re-identification. Li et al.~\cite{li2017deep} performed multi-task training that includes ID classification, attribute recognition, contrastive loss, and triplet loss. Tang el al.~\cite{tang2017multi} made use of hand-crafted features such as color and introduced a multi-modal metric learning method to fuse these features with deep features extracted using a neural network. Moreover, GANs are being increasingly adopted in vehicle re-identification. The main intuition is to transfer the query image to a domain that makes it more robust and efficient to compare with other images~\cite{BMVC2017_186,lou2019embedding}. Zhou et al.~\cite{BMVC2017_186} proposed the generation of vehicle images from different views to deal with cross-view re-identification.

Large-scale datasets for vehicle re-identification are still recent and with the emergence of intelligent transportation, more research is being developed to improve this field. While it does introduce certain challenges different from person re-identification, the ability to find a re-identification method that performs well on any object of interest is important. In this paper, we do not make use of characteristics specific to the object of interest or feature division and show the importance of confidence when training a re-id model with a cross-entropy loss.

\section{Problem Formulation}\label{sssec:problem}

\begin{figure}[t]
\begin{center}
\includegraphics[width=\linewidth]{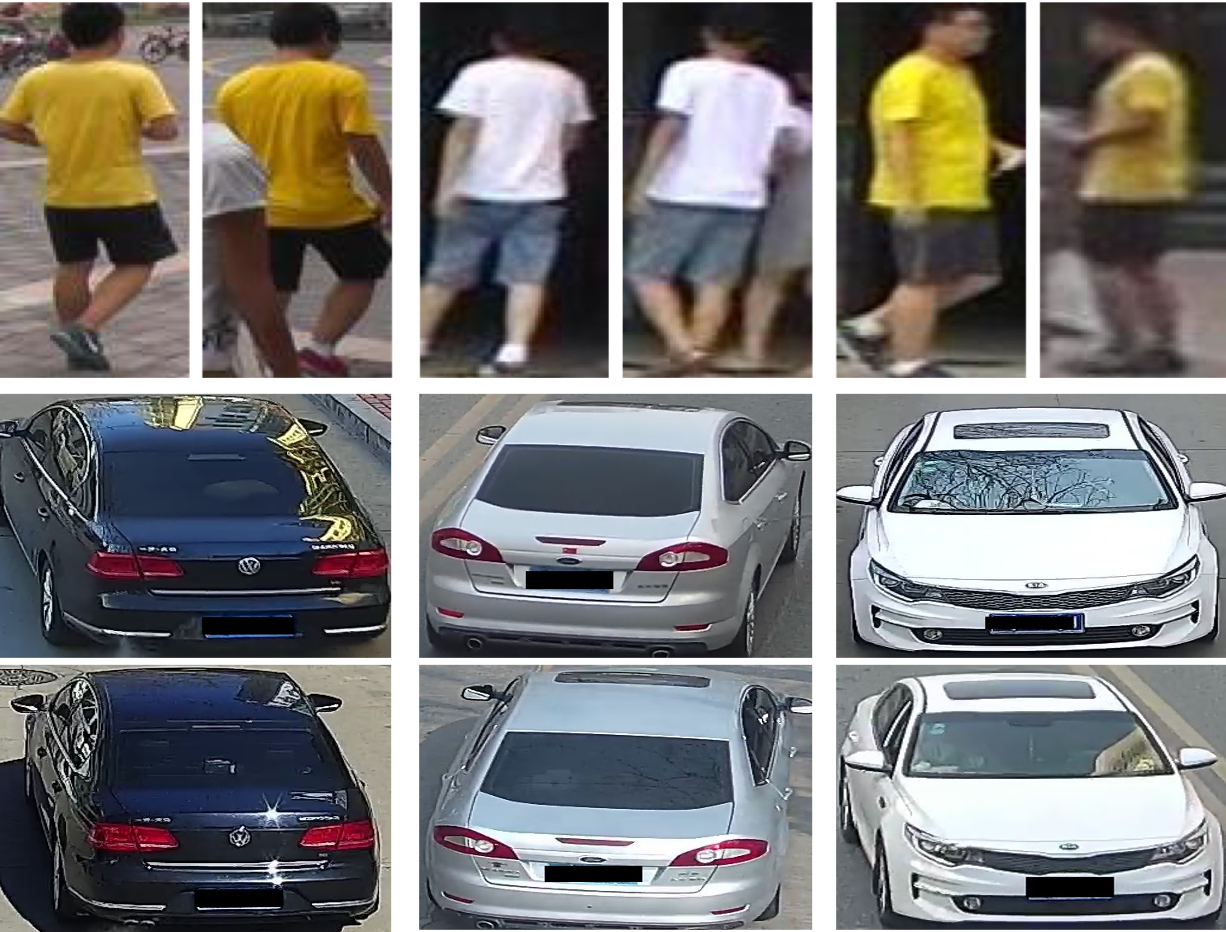}
\end{center}
\caption{Pairs of images of different IDs but very similar appearance. None of these images belong to the same agent. Images taken from Market1501~\cite{zheng2015scalable} \& VERI-Wild~\cite{Lou_2019_CVPR} datasets.}
\label{fig:differentID_similarPic}
\end{figure} %% TODO: Add image of vehicles that are similar to each other

A re-id model's main task is to distinguish between different agents across images. As previously stated, this is a challenging task since it tries to relate images of agents across different cameras as well as at different times. The fact that the images are captured under different circumstances might lead to subtle differences in hue and image color that can drastically effect the performance of a re-id model. Moreover, the illumination, background clutter, occlusion, and observable object parts are usually dramatically different which might easily fool the network and render it unusable. Even images captured by the same camera can have many of these variations.

Due to the challenges explained above, there is not always a clear margin of separation between individual agents. Pedestrians or vehicles in some cases have very subtle differences that separate them from each other making the task of identifying them even more challenging for an observer. A good example is shown in Figure~\ref{fig:differentID_similarPic} which introduces the inherent challenge we are trying to tackle in this paper.

The pedestrians within the images in Figure~\ref{fig:differentID_similarPic} are very difficult to discern from one another even for a human eye. Each pair of images show two different pedestrians who share very similar appearances. When a model is trained to separate these images, it might face difficulties doing so. Since the images are very similar and a network's only main goal is to reduce its loss, it will learn to focus on the pose or even the illumination of the images to discern them. These two variations are some of the many variations that previous methods try to overcome. This problem is also shared with vehicle re-identification. Since different vehicles might share the same model and color, many variations such as illumination, viewing angle, and weather might be used by the network to separate them.

Current state-of-the-art re-id systems train their own models by using the cross-entropy loss function. The cross-entropy calculates the number of bits needed for an event, which in this case is the label given the input, using the estimated probability distribution instead of the true distribution. In the case of training a neural network, the cross-entropy is minimized so that the model distribution is the same as that of the ground-truth, which is usually a one-hot encoding. This means minimizing this loss pushes the distribution of the model to output a high probability for the correct label while outputting very low probabilities for the others. The fact that cross-entropy requires that the logits for the ground-truth label to be much bigger than other labels pushes the network to take into consideration certain destructive variations to separate the different classes and especially for images such as in Figure~\ref{fig:differentID_similarPic}.

In order to modify the cross entropy in a way that solves the problem described above, we add a missing term to the loss function which allows it to not be confident about certain data points. Thus, the modified cross entropy loss function allows the network not to overfit on variations that are destructive for the re-id task and accept the fact that pedestrians or vehicles do sometimes look very similar. The idea of preventing the network from being very confident is not a new concept. However, its evaluation on other computer vision tasks, such as object detection, only leads to slight improvements in performance. From the reasoning based on Figure~\ref{fig:differentID_similarPic} as well as the characteristics of person and vehicle re-id datasets, we show in this paper that this concept, if applied to a simple baseline, can improve the results drastically and even outperform certain highly specialized state-of-the-art methods.

\section{Method}

\begin{figure*}[h]
\begin{center}
\includegraphics[width=\textwidth,height=6.5cm]{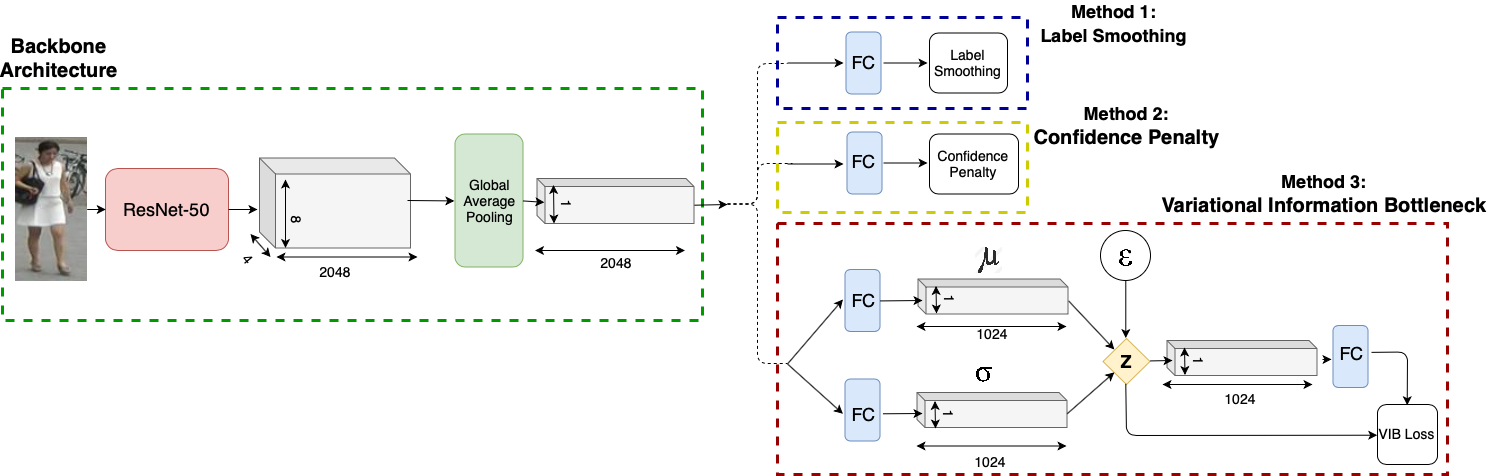}
\end{center}
\caption{Network architecture including the three methods being studied: Label Smoothing (LS), Confidence Penalty (CP), Variational Information Bottleneck (VIB). $\epsilon$ is the Gaussian noise needed for the reparameterization trick, $\mu$ and $\sigma$ are the mean and standard deviation respectively of the latent $Z$ distribution.}
\label{fig:VIB_arch}
\end{figure*}

Current re-id models face difficulties in distinguishing between different agents who share some visual similarities due to the model's objective of maximizing its confidence in its predictions, as previously discussed. In this section, we place the three common methods into a common framework of a cross-entropy term and a KL divergence term. This will allow us to investigate common properties and to identify their differences.
We show how the three methods, LS, CP, and VIB, are specific instances of our common framework which forces the model to be less confident of its predictions than a plain cross entropy term. These methods usually show a small improvement when used during training in other computer vision tasks~\cite{Szegedy_2016_CVPR,DBLP:journals/corr/PereyraTCKH17, DBLP:journals/corr/AlemiFD016}; however, we show that, because of the problems specified in Section~\ref{sssec:problem}, these methods provide a drastic boost for the task of person and vehicle re-id.

%\subsection{Penalizing Confidence}

% THE FOLLOWING PARAGRAPH DOES NOT INCLUDE VIB SCENARIO
% A network is confident of its predictions when it has a low output entropy. In order to penalize these confident predictions, the output entropy should be increased. Heuristically, the distribution with the maximum entropy corresponds to the least amount of knowledge which we take to be the uniform distribution over our flat categories. Thus, we add a KL divergence between the output distribution and a uniform distribution to the cross-entropy loss. This can be done in two ways by using either forward KL or backward KL. Each has different effects on the  dynamics of the training procedure. The general formulation then would be:

% \begin{equation}
% L = \alpha L_{cross} + \beta KL[A,B].
% \label{eq:generalEq_KL}
% \end{equation}

We will bring all methods into the following common framework of a cross-entropy regularized with a KL divergence where the loss $L$ is
\begin{equation}
    L = \alpha H(q,p) + \beta KL
    \label{eq:generalEq_KL}
\end{equation}
where $q=q(y|x)$ is the ground truth distribution of the output id $y$ given the input image $x$, $p=p(y|x)$ is the predicted
distribution, $H(q,p)$ is the cross-entropy between $q$ and $p$ and $KL$ is the Kulback-Leibler divergence~\cite{kullback1951information}.
In ReID, we have multiple ids $y \in Y$ where we denote the number of $Y$ -- the number of classes -- as $C$.

\subsection{Label Smoothing}

With Label Smoothing~\cite{Szegedy_2016_CVPR}, the predicted distribution $p$ is regularized towards the uniform distribution~$u$ with the KL divergence term:

\begin{equation}
    L_{LS} = \alpha H(q,p) + \beta KL[u, p] \;\;\;.
    \label{eq:label_smooth_KL_1}
\end{equation}

In this form, the KL divergence is the expectation over the uniform distribution of the logarithmic difference between $u$ and $p$. Forming the convex mixture with $\alpha \equiv 1-\beta$ and expanding this equation yields:
\begin{align}
    L_{LS} &= -(1-\beta)\sum_{y \in Y} q \log p
              + \beta \sum_{y \in Y} u \log \frac{u}{p} \\
           &= -\sum_{y \in Y}\left[
              \beta u + (1-\beta) q \right]\log p \\
           &= H(q_{LS}, p)
    \label{eq:label_smooth_KL_exp}
\end{align}
where we dropped the constant term $u \log u$.
We arrived at a single cross entropy without a KL divergence
and defined a new label-smoothed ground truth distribution $q_{LS}$.
The uniform distribution $u$ is over the $C$ classes, \textit{i.e.}, it evaluates to $1/C$. For a sample $n$:
\begin{equation}
    q_{LS}(y|x_n) =
    \begin{cases}
        1 - \frac{(C-1)\beta}{C}   & \textrm{true} \, y,\\
        \frac{\beta}{C}            & \textrm{otherwise}.
    \end{cases}
\label{eq:label_smoothing}
\end{equation}

From a KL divergence between $u$ and $p$, we obtained a loss function similar to the regularizer introduced by Szegedy et al.~\cite{Szegedy_2016_CVPR}, which aims at allowing a model to be less confident about its prediction. It regularizes a softmax classifier by assigning a small value to all ground-truth labels.

This method makes sure that the label for the correct class does not become much larger than all other classes and thus prevents the network from over-fitting. When label smoothing was proposed and tested on ImageNet, it showed a small improvement of around 0.2\% for top-1 error. Even though it did not show a huge improvement, we show in Section~\ref{sssec:results} that this method has a bigger effect on the task at hand based on the arguments stated in Section~\ref{sssec:problem}.
% As can be observed in Figure~\ref{fig:VIB_arch}, this method requires only modification to the cross-entropy loss function where the modified ground-truth distribution $q_{LS}$ is used.

\subsection {Confidence Penalty}

Reversing the arguments of the KL divergence, we obtain an equation for confidence penalty~\cite{DBLP:journals/corr/PereyraTCKH17}:

\begin{equation}
    L_{CP} = \alpha H(q,p) + \beta KL[p,u] \\
    \label{eq:KL_CP}
\end{equation}

Comparing Eq.~\ref{eq:KL_CP} to Eq.~\ref{eq:label_smooth_KL_1}, we can observe the main difference. The error calculation, in this case, between the uniform and predicted distribution is weighted by the predicted distribution. Expanding this equation:
\begin{align}
    L_{CP} &= \alpha H(q,p) + \beta \sum_{y \in Y} p \log \frac{p}{u} \\
    % &= \alpha H(q,p) + \beta \sum_{y \in Y} p\log(C\times p) \\
    % &= \alpha H(q,p) + \beta \sum_{y \in Y} p\log(p) + \log(C) \\
    &= \alpha H(q,p) - \beta H(p)
    \label{eq:KL_CP_exp}
\end{align}
where $\sum_{y \in Y} p \log u$ is removed since it is a constant.
We notice that the resulting loss function aims at maximizing the entropy of the predicted distribution. The increase in entropy forces the network to be less certain of its predictions. Pereyra et al.~\cite{DBLP:journals/corr/PereyraTCKH17} reached a similar conclusion and showed that, by applying this method, they got a smoother predicted distribution as well as a small improvement on MNIST. This method, however, did not show an improvement on a more difficult dataset such as CIFAR-10.
Similar to label smoothing, this method does not require any architecture modification as shown in Figure~\ref{fig:VIB_arch}.

\subsection{Deep Variational Information Bottleneck}

Another way to increase the entropy of the output distribution is to force latent representations to be similar to each other irrespective of the input. This means that information distinguishing different samples is being lost. This idea can be derived from the information bottleneck (IB) principle~\cite{Tishby99theinformation} where the mutual information between the input and latent representations is minimized. The IB principle~\cite{Tishby99theinformation} is a technique that tries to find the best trade-off between accuracy and complexity of latent variables. Latent variables are hidden variables that describe a specific input while maintaining all the relevant information needed for a specific task. The information bottleneck method tries to maximize this objective:
\begin{equation}
  \max_{p(z|x)}  \alpha I(z;y) - \beta I(z;x)
  \label{eq:IB}
\end{equation}
where $z$ is the latent variable. Based on the above equation, the objective is to learn a representation $z$ that is very informative about $y$ while compressive about $x$. In order to apply the IB objective to a neural network, Alemi et al.~\cite{DBLP:journals/corr/AlemiFD016} approximated a lower bound to the information bottleneck by using variational inference and the reparameterization trick introduced by Kingma et al.~\cite{Kingma2014AutoEncodingVB} to introduce a new objective function referred to as Variational Information Bottleneck~(VIB).

When applying this method, the model is divided into an encoder that takes the input $x$ and maps it to a distribution describing the latent space $z$. The encoder outputs both the mean $\mu$ and standard deviation $\sigma$ that describe this distribution. Then the predicted latent distribution is used to sample a specific latent representation. To force the second part of equation~\ref{eq:IB} to be maximized, this distribution should not depend on the input thus forcing the representation $z$ to forget some information about it. This is done by minimizing the divergence between the encoder's distribution $w = w(z|x)$ and the prior $r$ which is an isotropic multivariate Gaussian $r(z)$. The resulting objective function to minimize is:
\begin{equation}
    L_{VIB} = \alpha H(q, \tilde{p}) + \beta KL[w, r]
    \label{eq:VIB}
\end{equation}
where $\tilde{p} = p(y|f(x, \epsilon))$.

In order to compute the KL divergence analytically and back-propagate using its gradients, $w$ is approximated by a multivariate Gaussian distribution with a diagonal covariance matrix. As can be seen in equation~\ref{eq:VIB}, if $\beta\to\infty$, the latent representation would follow a distribution independent of the input. This is somewhat similar to the effect of both confidence penalty and label smoothing where a single representation is forced to contain some information about more than one label. However, VIB applies this restriction directly to the latent space. Using this method while training, Alemi et al.~\cite{DBLP:journals/corr/AlemiFD016} showed close results to state-of-the-art models while using less information about the input which is measured using mutual information $I(x;z)$.

Compared to previously mentioned methods, in order to use the VIB loss, a fully connected layer is added at the output of the ResNet-50 base model to compute the mean and standard deviation as shown in Figure~\ref{fig:VIB_arch}.

\section{Discussion}

% LS vs CP, they both work on output, KL divergence reversed, error of how far the output is from the uniform is weighted by either u=1/K or p(y|x) for LS or CP respectively
% If r(Z) was uniform , all methods would look similar except for the VIB->latent while LS and CP-> output
As can be seen in the equations above, all three methods try to increase the uncertainty of the model or in other words, decrease its confidence.
% If the prior $r(Z)$ was chosen to be a multivariate uniform distribution, the methods would look very similar.
On one hand, label smoothing and confidence penalty act on the output distribution while on the other hand, VIB acts on the latent representation directly. Thus this requires the original architecture to be modified to accommodate the VIB loss.

In addition, when expressed in terms of KL divergence, both label smoothing and confidence penalty are very similar except for the fact that the KL divergences are reversed. The forward KL divergence uses a constant represented by $u = \frac{1}{C}$ (Equation~\ref{eq:label_smooth_KL_1}) to weight the log expression. However, the reverse KL divergence weighs using the output prediction of the model $p(y|x)$ which varies during the training process (Equation~\ref{eq:KL_CP}). In other words, confidence penalty does not equally penalize all the label predictions but implicitly gives more importance to predictions that the network incorrectly gives higher probabilities to and which are farther away than the uniform distribution. This weighing is adaptively changing as the network trains. Label smoothing however penalizes all label predictions equally. Moreover, label smoothing tries to prevent an output prediction of 0. This is because it is weighted by the uniform distribution $u$ which is always greater than zero. Confidence penalty, however, might force certain output predictions to be zero although $u$ is never zero.

By expressing all three methods in terms of KL divergence, we get more insight on why confidence penalty is able to outperform other methods. Compared to label smoothing, confidence penalty weighs the error by the network’s current prediction and thus is more adaptable to the different inputs. Moreover, VIB provides the network with less degree of freedom compared to confidence penalty. This is because the latter acts on the output distribution allowing the representation, which is the main feature used for ranking in re-id, to move more freely in the feature space.

\section{Experiments}

To evaluate our proposed method, we use publicly available person and vehicle re-identification datasets which are Market-1501~\cite{zheng2015scalable}, MSMT17~\cite{wei2018cvpr}, DukeMTMC-reID~\cite{ristani2016performance}, and VERI-Wild~\cite{Lou_2019_CVPR}.
\begin{description}[style=unboxed,leftmargin=0cm]
\item [{Market1501~\cite{zheng2015scalable}:}] The Market dataset is a well-known person re-identification dataset that contains 32,668 bounding boxes of 1,501 individuals captured using 6 cameras. These bounding boxes were obtained using the Deformable Part Model (DPM)~\cite{5255236}. The training set is made up of 751 identities with 12,936 images while the test set has 750 identities distinct from the one in the training set divided into query and gallery images.
\item [{MSMT17~\cite{wei2018cvpr}:}] This is a very recent dataset which was carried out over a long period of time. This benchmark contains a total of 126,441 bounding boxes of 4,101 identities captured using 15 cameras. The images vary in terms of location (outdoors, indoors), weather conditions (over a month), as well as different times of day (morning, noon, afternoon). The bounding boxes were obtained using Faster RCNN and corrected using labelers. Containing many variations makes this dataset challenging as well as a good benchmark to use.

\item [{DukeMTMC-reID~\cite{ristani2016performance}:}] The DukeMTMC-reID dataset is a small part of the bigger DukeMTMC dataset that is usually used for multi-target multi-camera tracking. It is taken from 8 different cameras, and the person bounding box is manually labeled. It is made up of 1,404 different identities with 702 identities used for training and 702 other identities used for testing.

\item [{VERI-Wild~\cite{Lou_2019_CVPR}:}] The VERI-Wild dataset is a recently released large-scale vehicle re-identification dataset with 416, 314 vehicle images of 40, 671 IDs captured by 174 cameras. Vehicle recording is unconstrained and thus contains vehicles from different angles. This makes this dataset very challenging. The test set is divided into three subsets: small, medium, and large. We show our results on the small and medium subset since the large subset requires more memory.
\begin{figure}[h]
\begin{center}
\includegraphics[width=\linewidth]{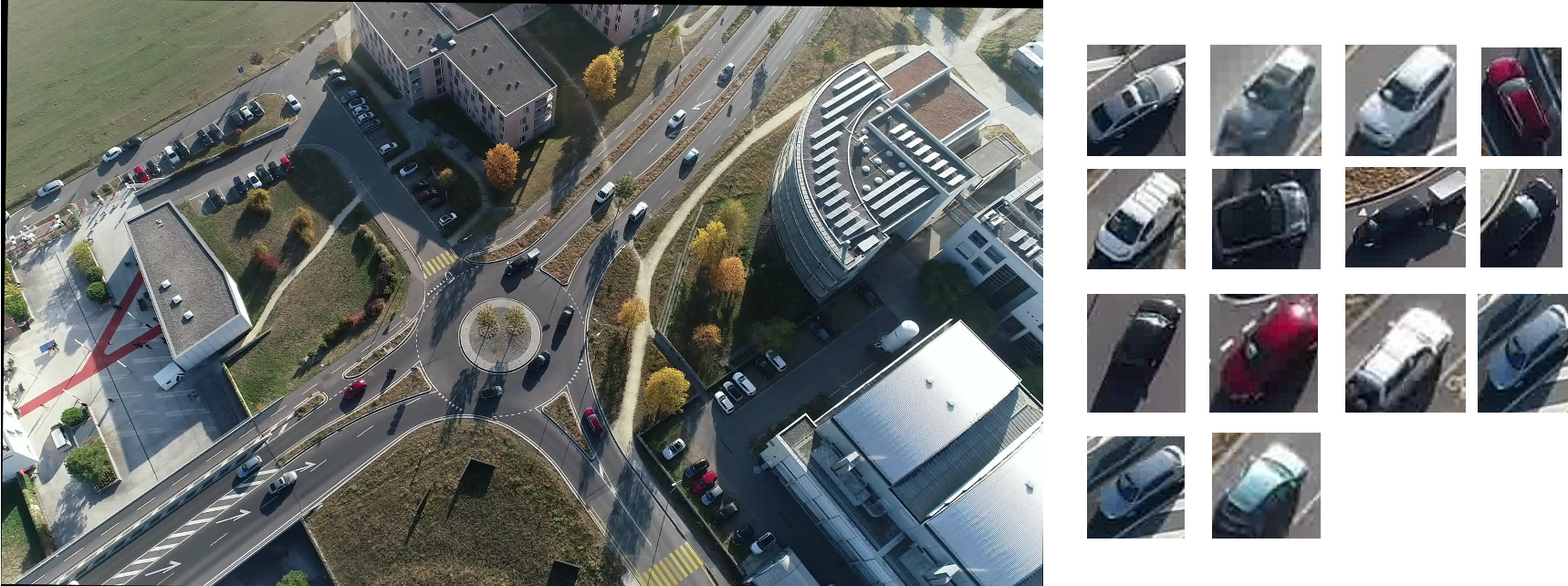}
\end{center}
\caption{An example from the EPFL Roundabout dataset and the vehicles that are detected. We make use of the crop of vehicles to build a re-identification dataset similar to VERI-Wild.}
\label{fig:epfl_roundabout}
\end{figure}
\item[{EPFL Roundabout:}] In order to validate the use of our methods on images recorded from drone view, we build the EPFL Roundabout dataset by recording vehicle flow of five different locations using a drone. The division of the dataset is similar to VERI-Wild. To encourage efficient training,  we set the training and testing ratio as 1:3. As a result, we obtain 85,268 images with 3,479 IDS. An image example is shown in Figure~\ref{fig:epfl_roundabout} The dataset will be made publicly available.

\end{description}
\subsection{Evaluation Protocol}
For evaluation, we use the cumulative matching characteristic (CMC) and Mean Average Precision (mAP). These two metrics are the most popular evaluation metrics since re-identification systems should be able to output all the correct matches (mAP) in addition to having high accuracy at different ranks (CMC). During testing, for every query, there is a list of gallery images ordered in increasing order according to their L2 distance from this query.

\subsection{Implementation Details}
\begin{table}[t]
\begin{center}
\begin{tabular}{|c|c|c|c|c|}
\hline
Parameters & Market-1501& DukeMTMC& MSMT17&VERI-Wild\\
\hline\hline
$\text{LR}_{cross},\; \alpha$  &
$5\times 10^{-4},\;  2$ &
$2\times 10^{-4}, \; 1$ &
$3\times 10^{-4},\; 4$ &
$1\times 10^{-3},\; 6$ \\
$\text{LR}_{LS},\; \alpha $   &
$5\times 10^{-4},\;  2$ &
$5\times 10^{-4}, \; 5$ &
$3\times 10^{-4}, \; 5$ &
$1\times 10^{-3},\; 6$ \\
$\text{LR}_{CP},\; \alpha$ &
$6\times 10^{-4},\; 3$ &
$6\times 10^{-4},\; 3$ &
$4\times 10^{-4},\; 5$ &
$1\times 10^{-3},\; 6$ \\
$\text{LR}_{VIB},\; \alpha$ &
$4\times 10^{-4},\; 6$ &
$6\times 10^{-4},\; 3$ &
$5\times 10^{-4},\; 6$ &
$1\times 10^{-3},\; 6$\\
$\beta_{CP}$ &  0.085 & 0.085&0.085 & 0.6\\
$\beta_{VIB}$&  0.01 & 0.01&0.01&0.01\\
\hline
\end{tabular}
\end{center}
\caption{Hyperparameters for the different datasets and methods. LR: learning rate, $\beta$: pre-factors for loss constraint, $\alpha$: pre-factor for cross-entropy}
\label{table:hyperparameters}
\end{table}
 The model was pre-trained using ImageNet. We do not add any layer to ResNet-50 when training both using label smoothing and confidence penalty except for a fully connected layer that outputs the different labels. When training the VIB algorithm, a fully-connected layer was added before the classification layer to output the mean and standard deviation which describe the distribution of the latent representations. A latent variable is then sampled from the predicted distribution. For all methods and datasets, hyperparameter tuning was performed for ResNet-50 in order to get the best possible accuracy.

\begin{description}[style=unboxed,leftmargin=0cm]

\item[Data Augmentation.]
We follow methods of data augmentations that are commonly used in the field of person re-identification. Since Market1501 uses DPM to obtain the bounding boxes, the images are initially randomly cropped. For all datasets, the inputs are resized to 256x128. Before providing them to the network, a random rectangle, with pixel values randomly assigned between [0, 255], is erased~\cite{DBLP:journals/corr/abs-1708-04896} from the images, and the resulting images are flipped horizontally with a probability of 0.5. This makes the network more robust to the orientation of the agents in the image as well as occlusion. Each image is then normalized and standardized using the mean and standard deviation provided when using a model pretrained on ImageNet. These transformations were applied only for the training set.

\item[Hyperparameter Tuning.]
Since the hyperparameters (e.g., learning rate, $\beta$, and $\alpha$) we are trying to optimize have multiplicative effects on the training procedure, the best method is to perform a log-space search. This is due to two reasons. The parameter is not too sensitive such that there may not be too much difference with 10 and 15 compared to 10 and 1000. The other reason is that using logarithmic scales allows us to search over a bigger space quickly.
\item[Training Procedure.]
The samples used to form the training batch are randomly sampled from the datasets. It does not require any special sampling such as the PK Sampling required by triplet loss\cite{Schroff_2015_CVPR}, which randomly samples P identities and then randomly K images for each identity to form a batch. The mini-batch has a size of 32 images. The model is trained for 300 epochs using AMSGrad~\cite{j.2018on} for all datasets with the learning rate decaying by 10 at epoch 20 and 40. In order to make sure that all models were trained with the best parameters, we perform hyperparameter tuning, as discussed previously. The different hyperparameters for the different datasets are shown in Table~\ref{table:hyperparameters}.
\item[Evaluation Procedure.]
For testing, the features that are extracted just before the last classification layer are used for the ranking process. The features for the queries and galleries are extracted and then compared to rank the gallery images relative to each query image. This is done when label smoothing or confidence penalty is used. When using the VIB loss, the network has an additional fully connected layer that outputs the mean and standard deviation for every latent dimension and a reparameterization trick that depends on random Gaussian noise. For ranking, we use the mean produced by the model as features for each image since this represents the average of the distribution over which the input image is mapped to. This is also due to the fact that the standard deviations tend to 1. To the best of our knowledge, using a latent representation sampled from a Gaussian parametrized by the predicted mean and standard deviation has not been tackled before for the person ad vehicle re-id task.

\end{description}

\section{Results}\label{sssec:results}

In order to show both qualitative and quantitative results, we split our results into three parts. In Sections~\ref{sssec:baselines} and \ref{sssec:state}, we compare our proposed methods to published baseline results and state-of-the-art methods respectively. In Section~\ref{sssec:effect}, we investigate the effect of the three methods on the ranking process of person and vehicle re-id. Although these methods were tested on ResNet-50, other re-id models can benefit from their positive effect on the performance especially when dealing with visually similar pedestrians.

\subsection{Properly Trained Baseline}\label{sssec:baselines}
\begin{table}
\centering
\begin{tabular}{ |c||c|c||c|c|}
 \hline
 & \multicolumn{2}{|c|}{Market1501}& \multicolumn{2}{|c|}{DukeMTMC} \\
  \hline
 Model & mAP &Rank1& mAP &Rank1 \\
 \hline\hline
ResNet-50 \cite{Liu_2018_CVPR}& 47.78  & 73.90& 44.99  & 65.22 \\
ResNet-50 \cite{Shen_2018_CVPR}& 59.8  & 81.4 & 55.5  & 75.3 \\
ResNet-50 \cite{Sarfraz_2018_CVPR}&   59.8  & 82.6&   50.3  & 71.5   \\
ResNet-50 \cite{Chang_2018_CVPR} &66.0  & 84.3&48.6  & 71.6 \\
ResNet-50 \cite{Huang_2018_CVPR}& 66.95  & 84.42& 57.34  & 75.60 \\
ResNet-50 \cite{Kalayeh_2018_CVPR} &66.32 & 85.10 &54.77 & 73.70\\

\hline
Our ResNet-50&   70.2  & 87.5 &   59.6  & 78.6  \\
\hline
\end{tabular}
\vspace{0.5cm}
\caption{Comparison with published ResNet-50 results on the Market-1501 and DukeMTM-reID dataset.}
\label{table:baseline_market}
\end{table}

We compare our baseline to previously reported results of ResNet-50 on the Market-1501 and DukeMTMC-reID datasets. The published results reported in Table~\ref{table:baseline_market} correspond to pre-trained ResNet-50 that used the cross-entropy loss similar to our method. As can be observed in Table~\ref{table:baseline_market}, there is a clear difference between our result and the results reported in published papers as well as amongst the published results themselves. Our properly trained baseline, which consists of a ResNet-50 model trained using a normal cross-entropy loss, was able to outperform all previous baselines. This table represents one of the many pitfalls that occurs when training a model. This is shown by the fact that papers that make use of exactly the same baseline have different results. This is usually due to the hyperparameters chosen. Another pitfall is that to compare different baselines and losses, the same hyperparameters are set. This is somewhat unfair since different baselines and losses optimize different parameters and in different ways thus requiring distinct hyperparameters. This is why we employ different learning rates for different datasets and methods as shown in Table~\ref{table:hyperparameters}. As a result, we were able to achieve, using the baseline, around $\sim$~3\% increase in mAP and rank-1 for both datasets.

\subsection{Comparison with State-of-the-Art}\label{sssec:state}

\begin{table}[h]
\centering
\begin{tabular}{ |c||c|c||c|c|}
 \hline
 &\multicolumn{2}{|c|}{Market1501}&\multicolumn{2}{|c|}{DukeMTMC} \\
  \hline
 Model & mAP &Rank1& mAP &Rank1 \\
 \hline\hline
CamStyle (R)\cite{Zhong_2018_CVPR} &  {71.55}  &  {89.49} &  {57.61}  &  {78.32}\\
HAP2S\_E+Xent(R)\cite{Yu_2018_ECCV}&74.49&89.73 &62.62&79.08\\
DuATM(!R)\cite{Si_2018_CVPR}&75.22 & 89.96&63.14 & 81.46\\
MLFN (!R)\cite{Chang_2018_CVPR}& 74.3  & 90.0 &62.8  & 81.0 \\
Shen et al.(R)\cite{Shen_2018_CVPR}& 75.3  & 90.1 & 63.2  & 80.3\\
PSE(R)\cite{Sarfraz_2018_CVPR} +ECN  &  {84.0}   & {90.3}&  {79.8}   & {85.2}\\
DaRe(!R)\cite{Wang_2018_CVPR} +RR & 86.7  & {90.9} & 80.0  & {84.4}\\
$\text{SPReID}^{w/fg}$(!R)\cite{Kalayeh_2018_CVPR}*&78.66 & 90.97&65.66 & 81.73\\
HA-CNN (!R) \cite{Li_2018_CVPR} & 75.7  & 91.2& 63.8  & 80.5\\
DuATM(!R)\cite{Si_2018_CVPR}**&76.62 & 91.42&64.58 & 81.82\\
$\text{SPReID}^{comb}$(!R)\cite{Kalayeh_2018_CVPR}*&79.67 & 91.45&68.78 & 83.3\\
P-Aligned (!R)\cite{Suh_2018_ECCV}&79.6&91.7&69.3&84.4\\
 SGGNN(R)\cite{Shen_2018_ECCV}&82.8&92.3&68.2&81.1\\
Deep Group RW(R)\cite{Shen_2018_CVPR}&82.5 & 92.7&66.4 & 80.7\\
 Mancs(R)\cite{Wang_2018_ECCV}&82.3&93.1&71.8&84.9\\
DNN+CRF(R) \cite{Chen_2018_CVPR}&81.6&93.5&69.5&84.9\\
PCB+RPP \cite{Sun_2018_ECCV}& 81.6& \textbf{93.8}& 69.2& 83.3\\
 P-Aligned (!R)\cite{Suh_2018_ECCV}+RR&{89.9}&93.4&{83.9}&{88.3}\\
\hline
Our ResNet&   70.7  & 87.2 &   59.6  & 78.6  \\
  Our ResNet(VIB) &   76.1  & 90.2 &   62.4 & 80.7\\
  Our ResNet(LS) &   76.7  & 91.0&     {64.4}  &  {82.7}\\
  Our ResNet(CP) &78.2&91.4&    {66.8}  &  {83.9}\\
\hline
 Our ResNet+RR&   85.7  & 89.7 &   78.5  & 83.4  \\
 Our ResNet(VIB)+RR & 88.6&91.8&    79.0 & 84.3\\
 Our ResNet(LS)+RR&   89.1  & 92.2 &  {82.2}  &  {86.6}\\

 Our ResNet(CP)+RR & {90.0}&92.6&    {83.5}  &  {87.4}\\
 Our ResNet(VIB)+ECN & 88.2&92.0 & 78.9  & 85.1\\
 Our ResNet(LS)+ECN & 89.4&92.7& {83.2}  & {86.9}\\
 Our ResNet(CP)+ECN & \textbf{90.1}&93.1&   \textbf{84.1}  & \textbf{88.5}\\
  \hline
\end{tabular}
\vspace{0.5cm}
\caption{Comparison with state-of-the-art methods on Market-1501 and DukeMTMC-reID. (!R): uses model different than ResNet, (R): uses ResNet-50, ECN: Expanded Cross Neighborhood Re-Ranking\cite{Sarfraz_2018_CVPR}, "RR": k-reciprocal re-ranking\cite{Zhong_2017_CVPR}, Xent: Softmax, *: uses combination of 10 datasets for training, **: uses data augmentation during evaluation stage.}
\label{table:state_of_art_market}
\end{table}

\begin{table}[h]
\centering
\begin{tabular}{ |c||c|c|c|}
  \hline
\multicolumn{4}{|c|}{MSMT17} \\
  \hline
 Model & mAP &Rank1&Rank10\\
 \hline\hline
GoogleNet\cite{wei2018cvpr}&   23.0 & 47.6&71.8\\
PDC\cite{wei2018cvpr}&   29.7 & 58.0&79.4\\
GLAD\cite{wei2018cvpr}&   34.0 & 61.4&81.6\\
  \hline
  Our ResNet &   31.8  & 59.3   & 80.2\\
  Our ResNet(VIB)&   35.1  & 66.2   & 84.1\\
  Our ResNet(LS) &   36.9  & 66.8   & 84.9\\
  Our ResNet(CP) &   39.3  & 68.6   & 85.3\\
  \hline
 Our ResNet + RR &   49.8  & 65.7 & 79.8\\
  Our ResNet(VIB) +RR&   55.4  & 73.3   & 84.7\\
  Our ResNet(LS) + RR&   57.1  & 73.7   & 85.3\\
  Our ResNet(CP) + RR&   \textbf{59.1}  & \textbf{75.3}   & \textbf{85.8}\\
  \hline

\end{tabular}
\vspace{0.5cm}
\caption{Comparison with state-of-the-art on the MSMT17 dataset. }

\label{table:state_of_art_msmt17}
\end{table}

We evaluate our proposed confidence-based methods against recently published papers in person and vehicle re-id. Each of our methods is evaluated on five datasets: Market1501, DukeMTMC-reID, MSMT17, VERI-Wild, and EPFL Roundabout. We are able to reach state-of-the-art performance without any human-specific design and added complexity thus showing the importance of penalizing the confidence of a network in the task of re-identification. We also do not make use of data augmentation during the evaluation stage like DuATM~\cite{Si_2018_CVPR}.
\begin{description}[style=unboxed,leftmargin=0cm]
\item [{Evaluation on Market1501:}]
As shown in Table~\ref{table:state_of_art_market}, the models were able to reach state-of-the-art results. In order to better understand the importance of penalizing confidence compared to other methods, it is important to note some distinct differences. Confidence penalty was able to outperform HAP2S~\cite{Yu_2018_ECCV} which tried to deal with hard samples by giving them higher weights. Moreover, Mancs\cite{Wang_2018_ECCV}, which shows good performance, makes use of three different losses, attention layers, as well as a special sampling scheme. To compare our results with methods that include gallery-to-gallery information during inference, such as Deep Group RW~\cite{Shen_2018_CVPR} and SGGNN~\cite{Shen_2018_ECCV}, we apply re-ranking to our three methods. We were able to outperform these methods with a significant increase in mAP($\sim$ 7.5\%). As a result, we got state-of-the-art performance without the added complexity of learning new layers and parameters while tackling the problem stated in Section~\ref{sssec:problem}.

\item [{Evaluation on DukeMTMC-reID:}] Similar to the Market-1501 dataset, we achieved competitive results in all proposed methods with confidence penalty resulting in the best improvement (Table~\ref{table:state_of_art_market}). In addition to that, using Sarfraz et al.'s~\cite{Sarfraz_2018_CVPR} recent re-ranking method (ECN), we were able to get better results than PSE~\cite{Sarfraz_2018_CVPR} in both mAP and rank-1. It is important to note that SPReID augments the training data of both DukeMTMC-reID and Market1501 with 10 datasets resulting in a large number of training samples which would improve the performance of the network.

\item [{Evaluation on MSMT17:}] Since this is a bigger dataset with many variations, it proved to be a challenging benchmark\cite{wei2018cvpr}. Nonetheless, we were able to show a notable improvement over previous methods as well as over our own baseline (Table~\ref{table:state_of_art_msmt17}). Similarly, confidence penalty performed the best by achieving 68.6\% in rank-1 and 39.3\% in mAP. By applying re-ranking, both rank-1 and mAP are further improved to 75.3\% and 59.1\% respectively.
\iffalse
\begin{table}[!]
\centering
\begin{tabular}{|c||c|c||c|c|}
 \hline
 \multicolumn{5}{|c|}{VERI-Wild} \\
  \hline

 &\multicolumn{2}{|c|}{Small} &\multicolumn{2}{|c|}{Medium}\\
 \hline
 Model & mAP &Rank1 & mAP &Rank1\\
 \hline\hline
 GSTE~\cite{Bai2018} & 31.4 & 60.5& 26.18&52.12\\
 VERI-Wild~\cite{Lou_2019_CVPR}&  35.1& 64.0&  29.8& 57.8\\
\hline
\hline
  Our ResNet   & 47.2& 83.3&42.6&78.8\\ %/data/george-data/vehiclREID/resnet50_xent_veriWILD_lr0.0005_xent4_normalXent
  Our ResNet(LS) &   62.6& 88.9&57.2&85.5\\ % /data/george-data/vehiclREID/resnet50_xent_veriWILD_lr0.001_xent6_labelSmooth0.4/checkpoint_ep300.pth.tar
  %Our ResNet(CP) &   -- & --  & 65.2& 87.5\\ %/data/george-data/vehiclREID/resnet50_xent_veriWILD_confidencePenalty0.1_lr0.0005_xent2/
  Our ResNet(CP) &   70.7& 90.5&64.6&87.5\\%/data/george-data/vehiclREID/resnet50_xent_veriWILD_lr0.001_xent6_confidencePenalty0.6/checkpoint_ep300.pth.tar
  Our ResNet(VIB) &  \textbf{74.9}& \textbf{93.0}&\textbf{69.4}&\textbf{90.7}\\%/data/george-data/vehiclREID/resnet50vib_xent_veriwild_lr0.001_xent6_infoLoss0.01_MoreEpochs/checkpoint_ep100.pth.tar
\hline
 \end{tabular}
\vspace{0.5cm}
\caption{Comparison with state-of-the-art on VERI-Wild dataset.}
 \label{table:sota_veriwild}
 \end{table}
\fi

\begin{table}[!]
\centering
\begin{tabular}{|c||c|c||c|c|}
 \hline
 \multicolumn{5}{|c|}{VERI-Wild} \\
  \hline

 &\multicolumn{2}{|c|}{Small} &\multicolumn{2}{|c|}{Medium}\\
 \hline
 Model & mAP &Rank1 & mAP &Rank1\\
 \hline\hline
 GSTE~\cite{Bai2018} & 31.4 & 60.5& 26.18&52.12\\
 VERI-Wild~\cite{Lou_2019_CVPR}&  35.1& 64.0&  29.8& 57.8\\
\hline
\hline
  Our ResNet   & 45.7& 82.4&41.3&78.4\\ %10.91.1.26:/data/george-data/vehiclREID/resnet50_xent_veriwild_lr0.001_xent6_NoPenalty_changedOptimizerOrder/checkpoint_ep300.pth.tar
  Our ResNet(LS) &   57.7& 84.6&57.2&85.5\\ % /data/george-data/vehiclREID/resnet50_xent_veriwild_lr0.001_xent6_LabelSmooth_changedOptimizerOrder/checkpoint_ep300.pth.tar
  %Our ResNet(CP) &   -- & --  & 65.2& 87.5\\ %/data/george-data/vehiclREID/resnet50_xent_veriWILD_confidencePenalty0.1_lr0.0005_xent2/
  Our ResNet(CP) &   67.5& 90.2&61.8&87.0\\%10.91.1.26/data/george-data/vehiclREID/resnet50_xent_veriwild_lr0.001_xent6_ConfidencePenalty0.6_changedOptimizerOrder/checkpoint_ep300.pth.tar
  Our ResNet(VIB) &  \textbf{74.1}& \textbf{92.1}&\textbf{68.5}&\textbf{89.7}\\%10.91.1.26:/data/george-data/vehiclREID/resnet50vib_xent_veriwild_lr0.001_xent6_InfoLoss0.01_changedOptimizerOrder/checkpoint_ep300.pth.tar
\hline
 \end{tabular}
\vspace{0.5cm}
\caption{Comparison with state-of-the-art on VERI-Wild dataset.}
 \label{table:sota_veriwild}
 \end{table}

 \begin{table}
\centering
\begin{tabular}{ |c||c|c|}
 \hline
 &\multicolumn{2}{|c|}{EPFL Roundabout ReID}\\
  \hline
 Model & mAP &Rank1\\
 \hline\hline
 ResNet & 41.5  & 75.0 \\
 Our ResNet(LS) &55.9 & 84.4\\ % /data/george-data/vehiclREID/resnet50_xent_epflroundabout_lr0.001_xent6_LableSmooth0.4_changedOptimizerOrder/
 Our ResNet(CP) & 56.1 & 85.2\\ % /data/george-data/vehiclREID/resnet50_xent_epflroundabout_lr0.001_xent6_ConfidencePenalty0.1_changedOptimizerOrder/
 Our ResNet(VIB) &\textbf{52.7} & \textbf{82.5}\\
 \hline
\end{tabular}
\vspace{0.5cm}
\caption{Effect of our methods on EPFL Roundabout ReID}
\label{table:epflroundabout}
\end{table}

\item [{Evaluation on VERI-Wild:}] This dataset contains a large amount of images and is divided into small, medium, and large subsets. We evaluate our model on the small and medium subset and are able to achieve state-of-the-art performance (Table~\ref{table:sota_veriwild}). Compared to the person re-id datasets, using the VIB method achieves the best performance on VERI-Wild. This might be due to the fact that vehicles face the problems of similar appearance and different IDs more frequently. Thus, a more strict method of penalizing certainty is required compared to confidence penalty and label smoothing.

%\begin{table}
%\centering
%\begin{tabular}{ |c||c|c|}
% \hline
% &\multicolumn{2}{|c|}{EPFL Roundabout ReID}\\
%  \hline
% Model & mAP &Rank1\\
% \hline\hline
% ResNet & 27.7  & 55.3 \\
% Our ResNet(LS) &66.0& 70.4\\
% Our ResNet(CP) & 70.7& 74.4\\
% Our ResNet(VIB) &\textbf{57.2}&\textbf{84.2}\\
% \hline
%\end{tabular}
%\vspace{0.5cm}
%\caption{Effect of our methods on EPFL Roundabout ReID}
%\label{table:epflroundabout}
%\end{table}

\item [{Evaluation on EPFL Roundabout:}] Since the images are recorded from a drone view, vehicles are small and lack certain details. This makes it more challenging to discriminate between different vehicles and a a result suffers more from the problem discussed in Section~\ref{sssec:problem}. However, we were able to drastically improve the performance of the baseline by making use of confidence penalty, label smoothing, and VIB. Label smoothing, in this case shows, the biggest improvement of around 14\% and 10\% in both mAP and Rank1 respectively.(Table~\ref{table:epflroundabout}).

\end{description}

\subsection{Effect of Proposed Methods}\label{sssec:effect}
\begin{figure}[t]
\begin{center}
    \begin{subfigure}[h]{\linewidth}
        \centering
        \includegraphics[width=\linewidth,height=0.5\linewidth]{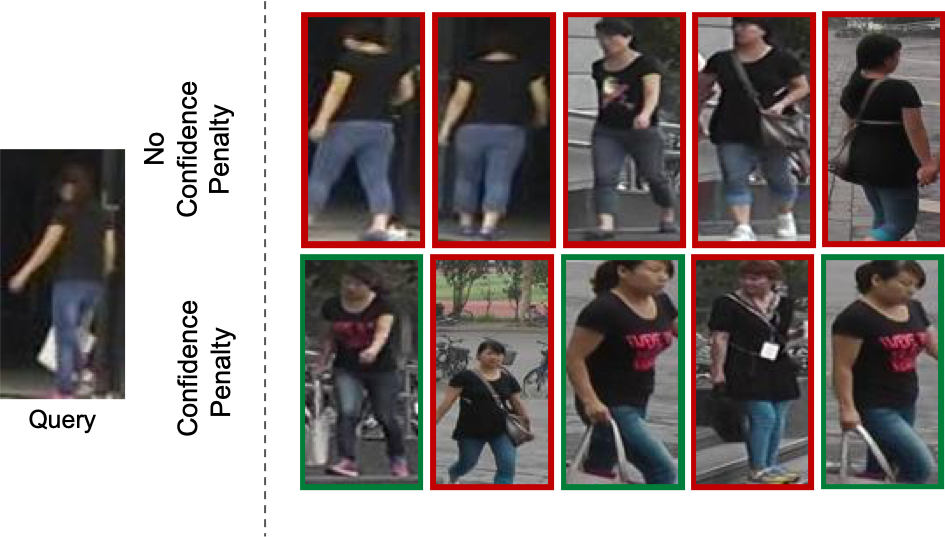}
    \end{subfigure}
    \begin{subfigure}[h]{\linewidth}
        \centering
        \includegraphics[width=\linewidth,height=0.5\linewidth]{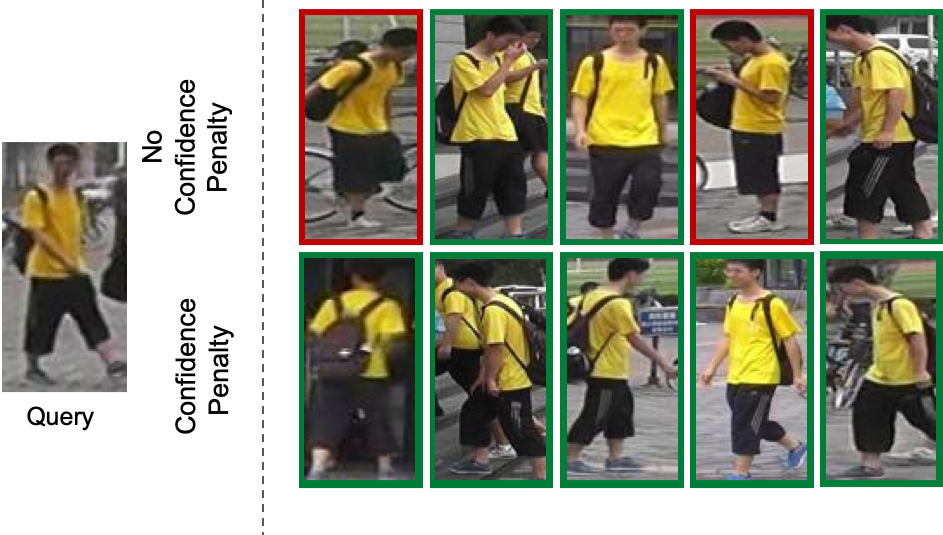}
    \end{subfigure}

\end{center}
\caption{Qualitative comparison of using confidence penalty on unseen Market1501 test samples. The gallery images are ranked according to L2 distance (top-5, left to right). Red frame indicates wrong ID while green frame indicates correct ID compared to the query. Best viewed in color.}
\label{fig:compare_CP_baseline}
\end{figure}

\begin{figure}[t]
\begin{center}
    \begin{subfigure}[h]{\linewidth}
        \centering
        \includegraphics[width=\linewidth]{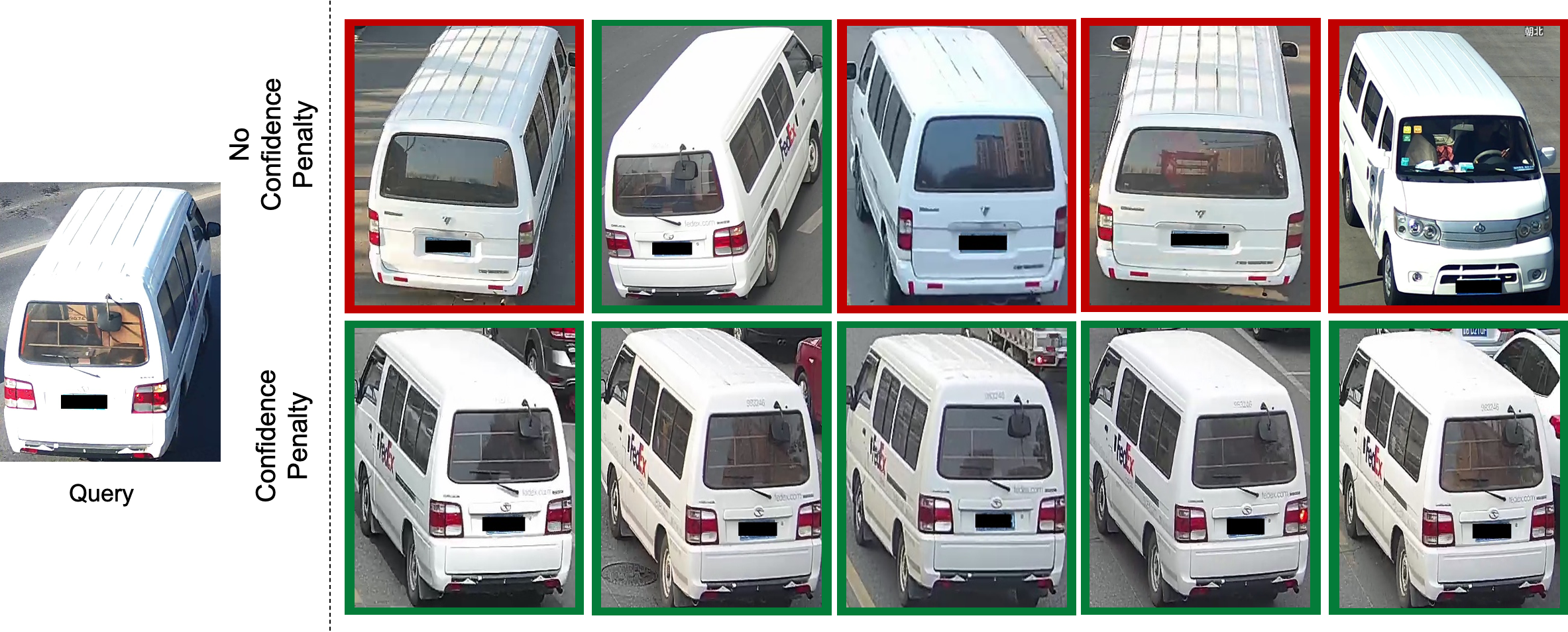}
    \end{subfigure}
    \begin{subfigure}[h]{\linewidth}
        \centering
        \includegraphics[width=\linewidth]{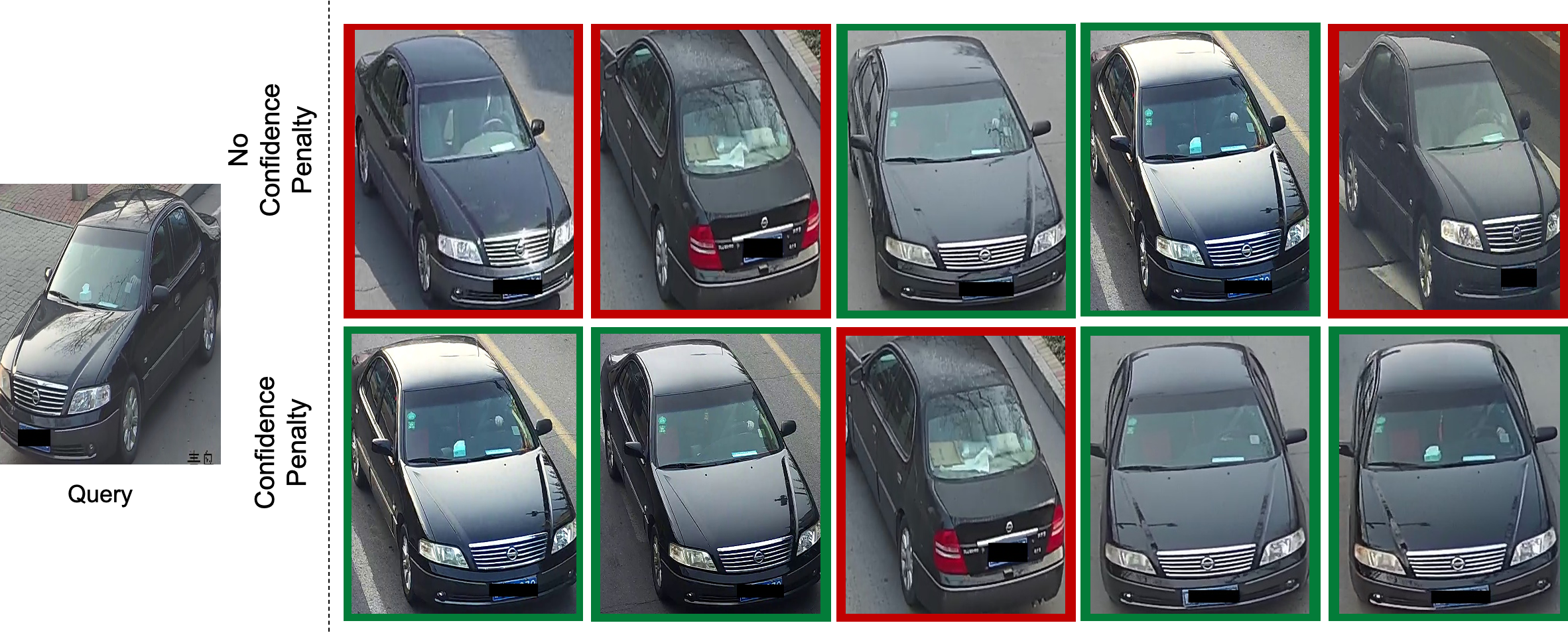}
    \end{subfigure}

\end{center}
\caption{Qualitative comparison of using confidence penalty on unseen VERI-Wild test samples. The gallery images are ranked according to L2 distance (top-5, left to right). Red frame indicates wrong ID while green frame indicates correct ID compared to the query. Best viewed in color.}
\label{fig:compare_CP_baseline_veriwild}
\end{figure}

In addition to achieving state-of-the-art performance, it is also important to understand the effect of these three methods on the ranking process. All three methods aim at allowing the network to share some representation among different classes. This prevents the network from focusing on undesirable information when separating very similar-looking pedestrians. To show this effect, we compare the confidence penalty model against the baseline model since it resulted in the best performance in person re-id. As can be seen, the test samples presented in Figure~\ref{fig:compare_CP_baseline} and  Figure~\ref{fig:compare_CP_baseline_veriwild} are difficult to rank even for an observer. This confirms the intrinsic difficulty of person and vehicle re-id stated in Section~\ref{sssec:problem}. When confidence penalty is not used for training, the network focuses on unimportant variations between the images. For instance, in both sets of samples (Figure~\ref{fig:compare_CP_baseline}), the incorrect gallery images are very similar to the query image despite belonging to a different person. The baseline links the query image to the gallery images by possibly focusing on the background, shirt color, posture, and body rotation of the pedestrian in question. The same applies for vehicles. These characteristics are typically features that can confuse the model leading to wrong identification. Adding the confidence penalty is observed to remedy this challenge, as can be seen for all test samples provided. Adding the confidence penalty helps the model capture the subtle differences between multiple pedestrians that the baseline tends to misidentify. These are ideal examples of why confidence penalty drastically improved re-identification compared to less significant improvements in other computer vision tasks.

\section{Ablation Study}

\subsection{Combination of Penalties}
\begin{table}[h]
\centering
\begin{tabular}{ |c||c|c||c|c|}
 \hline
 &\multicolumn{2}{|c|}{Market1501}&\multicolumn{2}{|c|}{DukeMTMC} \\
  \hline
 Model & mAP &Rank1& mAP &Rank1 \\
 \hline\hline
 ResNet &70.7  & 87.2 &   59.6  & 78.6\\
 Our ResNet(CP) &\textbf{78.2}&91.4&    \textbf{66.8}  &  \textbf{83.9}\\
 \hline
 Our ResNet(VIB+LS) &76.6& 90.4&63.9& 82.0\\
 Our ResNet(VIB+CP) &76.8& 90.4&62.0& 80.6\\
 Our ResNet(LS+CP) &76.7& \textbf{91.7}&63.6& 82.4\\
 Our ResNet(VIB+LS+CP) &76.9& 90.4&63.5& 81.2\\
\hline
\end{tabular}
\vspace{0.5cm}
\caption{Ablation study of different penalty combinations}
\label{table:combination_ablation}
\end{table}

Since the addition of the three methods to the baseline leads to an improvement in performance, one might wonder the effect of the different combination of these methods. As shown in Table~\ref{table:combination_ablation}, applying only confidence penalty (CP) leads to the best result. Intuitively, the combinations of the methods increases the restriction on the model preventing it from learning useful representations. Meanwhile,
adding these methods together still leads an improved performance compared to the baseline which again affirms the beneficial effect that they have.

\subsection{Penalties on State-of-the-Art methods}

\begin{table}[h]
\centering
\begin{tabular}{ |c||c|c||c|c|}
 \hline
 &\multicolumn{2}{|c|}{Market1501}&\multicolumn{2}{|c|}{DukeMTMC} \\
  \hline
 Model & mAP &Rank1& mAP &Rank1 \\
 \hline\hline

 %PCB &77.4& 92.3&66.1& 81.7\\
 %PCB+RPP &81.6& 93.8&69.2& 83.3\\
 PCB (own)\cite{Sun_2018_ECCV} &78.5& 92.4&69.6& 83.8\\
 Our PCB(LS) &77.1& 91.7&69.0& \textbf{84.6}\\
 Our PCB(CP)&\textbf{78.8}& \textbf{93.2}&\textbf{70.1}& 84.3\\
 \hline
 PCB+RPP (own)\cite{Sun_2018_ECCV} &81.5& 93.3&71.3& 84.4\\
 Our PCB+RPP(LS) &80.9& 92.8&71.3& \textbf{85.3}\\
 Our PCB+RPP(CP) &\textbf{81.6}& 93.3&\textbf{71.5}& 84.2\\
 \hline\hline
 HA-CNN\cite{Li_2018_CVPR} &75.7& 91.2&63.8& 80.5\\
 HA-CNN(own) & 78.6& 91.6&67.3&83.1\\
 \hline
 Our HA-CNN(LS) &80.1& 92.0&68.4& 83.6\\
 Our HA-CNN(CP)&\textbf{81.1}&\textbf{92.5}&\textbf{69.7}&\textbf{83.8}\\
 \hline
 HA-CNN(G)&72.3&87.9&60.0&78.7\\
 Our HA-CNN(CP+G)&\textbf{77.3}&\textbf{91.5}&\textbf{63.4}&\textbf{82.0}\\
 %HA-CNN(CP- only global(Norm))[16]&78.5&90.9&65.8&82.2\\
 \hline
 HA-CNN(L)&73.8&89.5&62.3&81.0\\
 Our HA-CNN(CP+L)&\textbf{75.5}&\textbf{91.2}&\textbf{62.8}&81.0\\
 %HA-CNN(CP- only local(Norm))[16]&77.0&91.4&65.5&81.7\\
 %\hline

 %HA-CNN(Xent-only global(Norm))[16]&73.6&88.5&61.8&78.6\\
 %HA-CNN(Xent-only local(norm))[16]&75.4&89.5&64.6&80.7\\
\hline
 \hline
BFE (Xent)\cite{Dai2018BatchFE}& 83.7& 93.5&73.5&86.4\\ %% MIGHT WANT TO REMOVE SINCE USES LS to get SoA
Our BFE (CP)&\textbf{85.7}&\textbf{94.2}&\textbf{76.1}&\textbf{88.6}\\
 \hline
\end{tabular}
\vspace{0.5cm}
\caption{Effect of CP and LS on PCB, HA-CNN and BFE. Xent: Cross-Entropy, G: using global features, L: using local features}
\label{table:pcb_hacnn}
\end{table}

To study the effect of our best method on the different state-of-the-art methods, we test confidence penalty on PCB~\cite{Sun_2018_ECCV}, HACNN~\cite{Li_2018_CVPR}, and BFE~\cite{Dai2018BatchFE}. It is important to note that PCB divides the representation into multiple features that are then used for identification. These features are referred to as local features since they are spatially local to a certain region of the input. In addition to local features, HACNN also extracts a global representation from the whole input. In comparison, BFE performs person re-identification using only a global representation. Table~\ref{table:pcb_hacnn} shows that the gain in performance for PCB is between +0.3 to +1.5\%. One intuition behind the limited gain is that PCB divides its features into multiple parts (local features) before applying global average pooling. This allows the representations to be fine-grained focusing on the details that differentiate visually similar inputs. To analyze this, we test CP on HA-CNN that uses global and local features. Its performance is improved ($\sim$+3\% mAP) to exceed PCB and to have on-par results with PCB+RPP. Also, CP has a bigger effect on global features than on local features in HA-CNN (Table~\ref{table:pcb_hacnn}) confirming our reasoning. The performance of another method, BFE, that uses only global features is also improved using CP compared to cross-entropy (Xent).

\subsection{Vehicle Ego-centric Pedestrian Re-Identification}

\begin{table}[!]
\centering
\begin{tabular}{ |c||c|c|}
 \hline
 &\multicolumn{2}{|c|}{nuScenes-ReID}\\
  \hline
 Model & mAP &Rank1\\
 \hline\hline
 ResNet &61.7  & 66.3 \\
 Our ResNet(LS) &66.0& 70.4\\
 Our ResNet(CP) &\textbf{70.7}&\textbf{74.4}\\
 Our ResNet(VIB) & 67.7& 70.4\\
 \hline
\end{tabular}
\vspace{0.5cm}
\caption{Effect of Confidence Penalty on NuScenes-ReID}
\label{table:nuscenes}
\end{table}
The dataset that were used to evaluate our methods are collected from cameras mounted in specific locations or using drones. In order to test the effectiveness of our method in a scenario where the camera is mounted on a car, we make use of the nuScenes dataset. This is the first dataset to contain all sensor information that a full autonomous vehicles requires from RGB camera, radars to lidars. Since the pedestrians are tracked across images and different cameras, we can build a re-identification dataset similar to Market1501. We call the resulting dataset, nuScenes-ReiD. This is a challenging dataset since images are recorded from the view of a moving car resulting in different pedestrian sizes. As shown in Table~\ref{table:nuscenes}, applying confidence penalty to the baseline significantly improves the performance (from 59.1\%mAP to 70.9\%). The code to build this re-id dataset will be made public.

\section{Conclusions}
An important task of transportation research is better analyzing and understanding traffic flow. Visual re-identification, the task of association similar agents, can aid in this goal. Thus, in this work we aim to deal with certain challenges that plague this task. First, we emphasize an intrinsic characteristic of person and vehicle re-identification that poses a problem to the network being trained. The classes that these re-id task try to separate are not as easy as separating cats and dogs. Different agents with different identities can have very similar appearances. We have demonstrated that three methods, that reduce a model's confidence, are able to deal with this problem while achieving state-of-the-art results. Confidence penalty proved to be the best and most lightweight amongst the three different methods. In addition, it is interesting to note that VIB is able to achieve similar results while using smaller representations. Both label smoothing and confidence penalty use a representation of dimension 2048 while VIB uses a representation of dimension 1024. These three methods can be leveraged to improve the performance of previous re-id methods as well. It remains an exciting future work to study their effect on other image retrieval and clustering tasks.

With the ability to identify pedestrians and cars across both time and space, this allows us to better understand how they move from one place to the other without going through the manual and expensive way of using surveys or interviews. Some methods have also been developed to estimate the OD matrices from existing observed traffic flows. These methods, however, require the collected data to be large and representative of the real distribution. This is where re-identification can play a major role. CCTV cameras already placed around entry and exit of different transportation stops can be used to associate agents that pass through them. For instance, a person can be detected entering a specific train and then this detection can be associated to the same person exiting at a different station and at different times. This task provides an automatic method of collecting data about passenger movements and thus allowing us to build an accurate OD matrix that can be used for different transportation tasks.

\bibliographystyle{unsrt}
%\bibliography{mybib,references}

\clearpage

\appendix
\title{Supplementary Material}
\maketitle
\section{Qualitative Results}

\begin{figure}[h]
\begin{center}
    \begin{subfigure}[h]{\linewidth}
        \centering
        \includegraphics[width=\linewidth,height=0.5\linewidth]{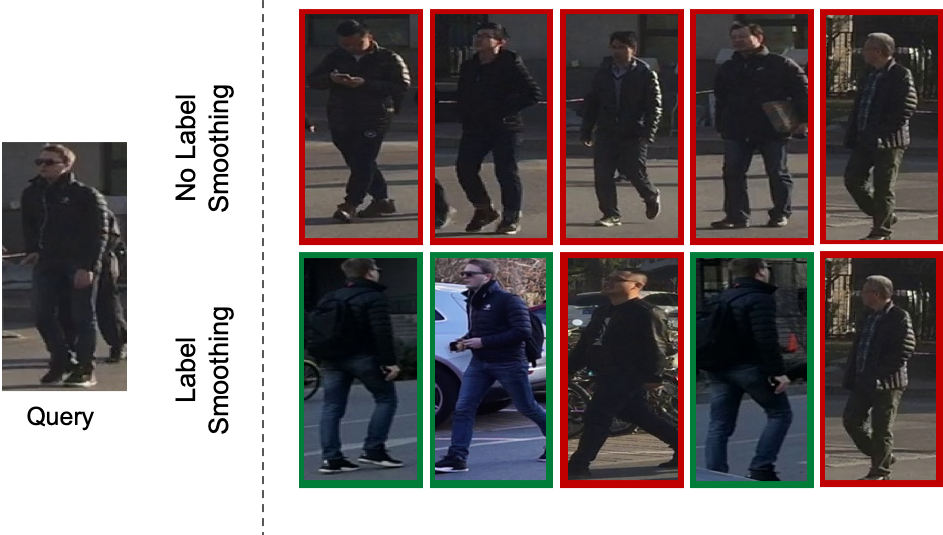}
    \end{subfigure}
    \begin{subfigure}[h]{\linewidth}
        \centering
        \includegraphics[width=\linewidth,height=0.5\linewidth]{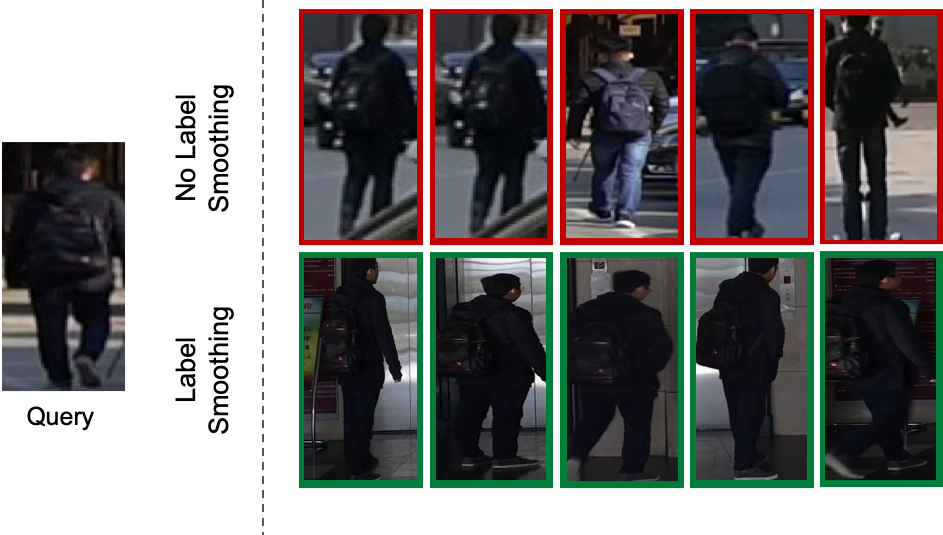}
    \end{subfigure}

\end{center}
\caption{Qualitative comparison of using label smoothing on unseen Market1501 test samples. The gallery images are ranked according to L2 distance (top-5, left to right). Red frame indicates wrong ID while green frame indicates correct ID compared to the query. Best viewed in color.}
% \label{fig:compare_CP_baseline}
\end{figure}
\begin{figure}[t]
\begin{center}
    \begin{subfigure}[h]{\linewidth}
        \centering
        \includegraphics[width=\linewidth,height=0.55\linewidth]{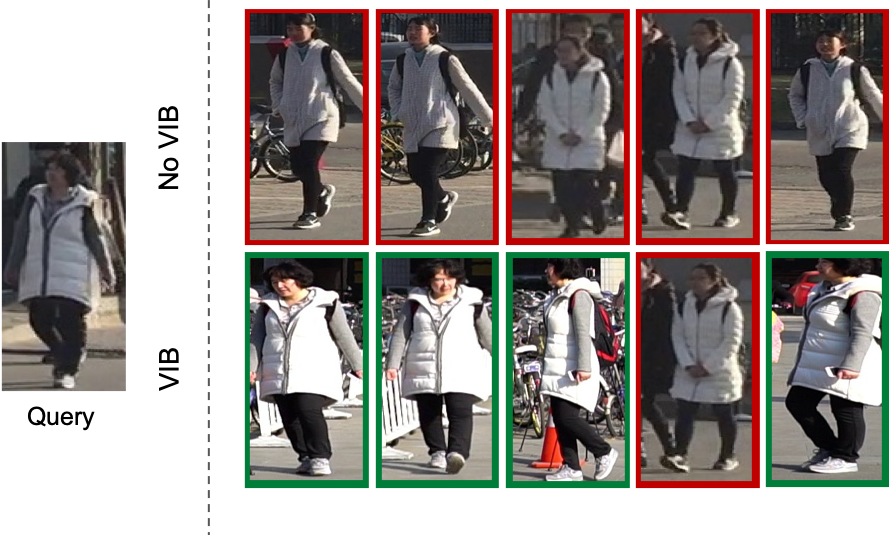}
    \end{subfigure}
    \begin{subfigure}[h]{\linewidth}
        \centering
        \includegraphics[width=\linewidth,height=0.55\linewidth]{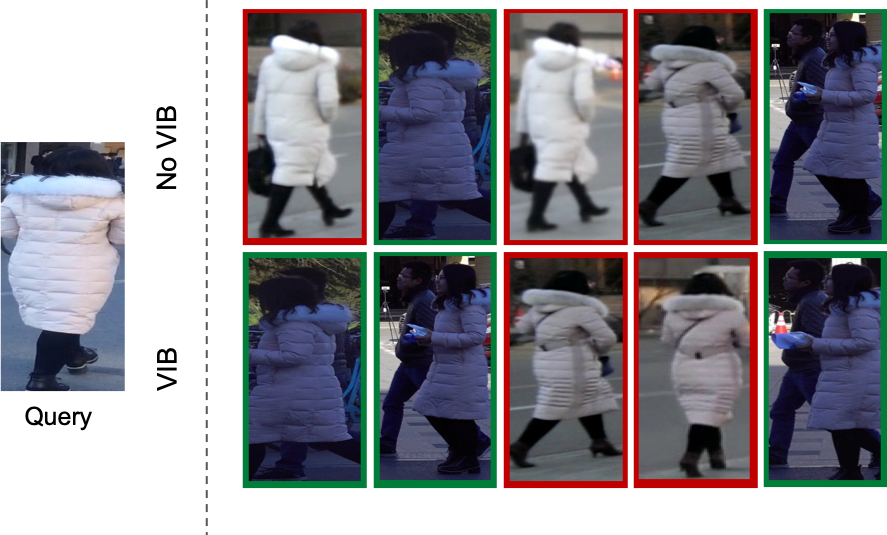}
    \end{subfigure}

\end{center}
\caption{Qualitative comparison of using VIB on unseen Market1501 test samples. The gallery images are ranked according to L2 distance (top-5, left to right). Red frame indicates wrong ID while green frame indicates correct ID compared to the query. Best viewed in color.}
% \label{fig:compare_CP_baseline}
\end{figure}
\begin{figure}
    \centering
    \includegraphics[width=1.05\linewidth,height=\linewidth]{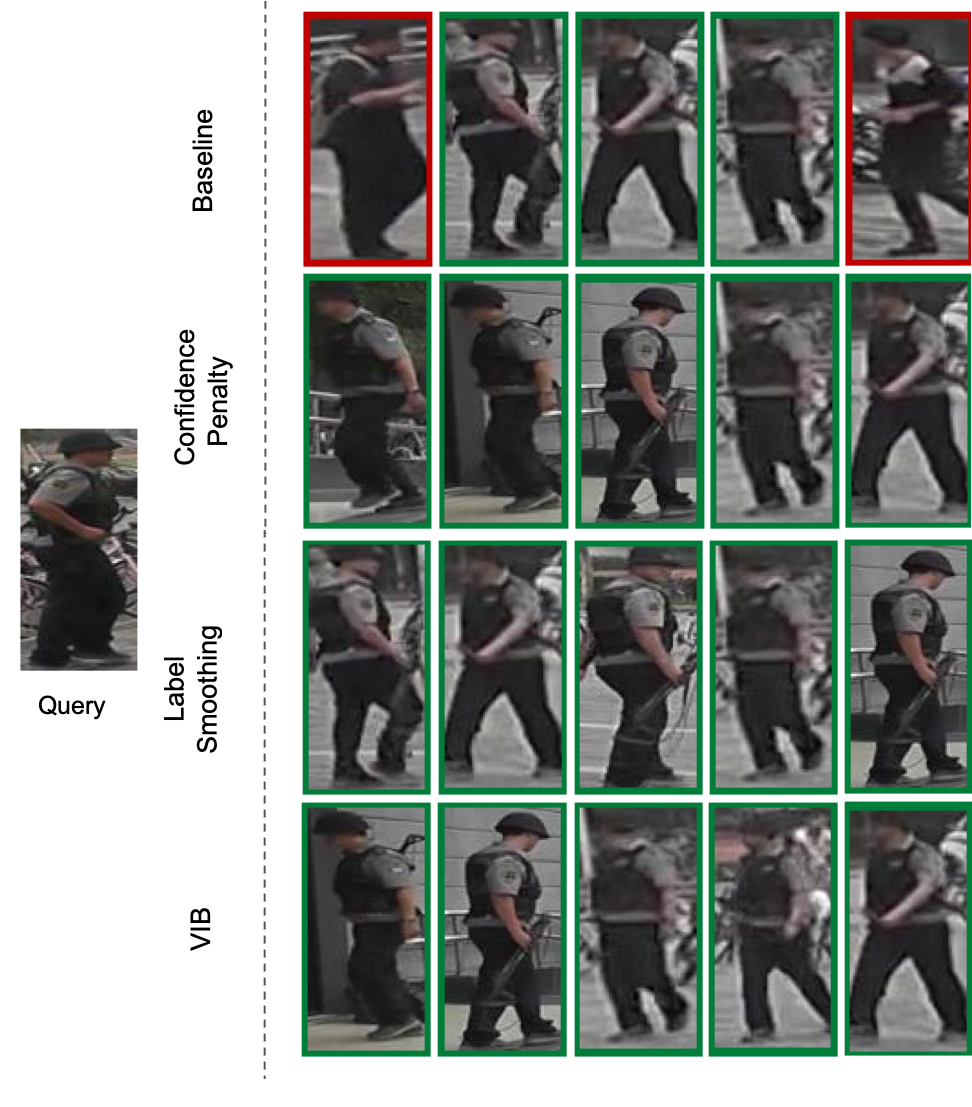}
    \caption{Qualitative comparison between the different methods on unseen Market1501 test samples. The gallery images are ranked according to L2 distance (top-5, left to right). Red frame indicates wrong ID while green frame indicates correct ID compared to the query. Best viewed in color.}
    % \label{fig:my_label}
\end{figure}
\begin{figure}
    \centering
    \includegraphics[width=1.05\linewidth,height=\linewidth]{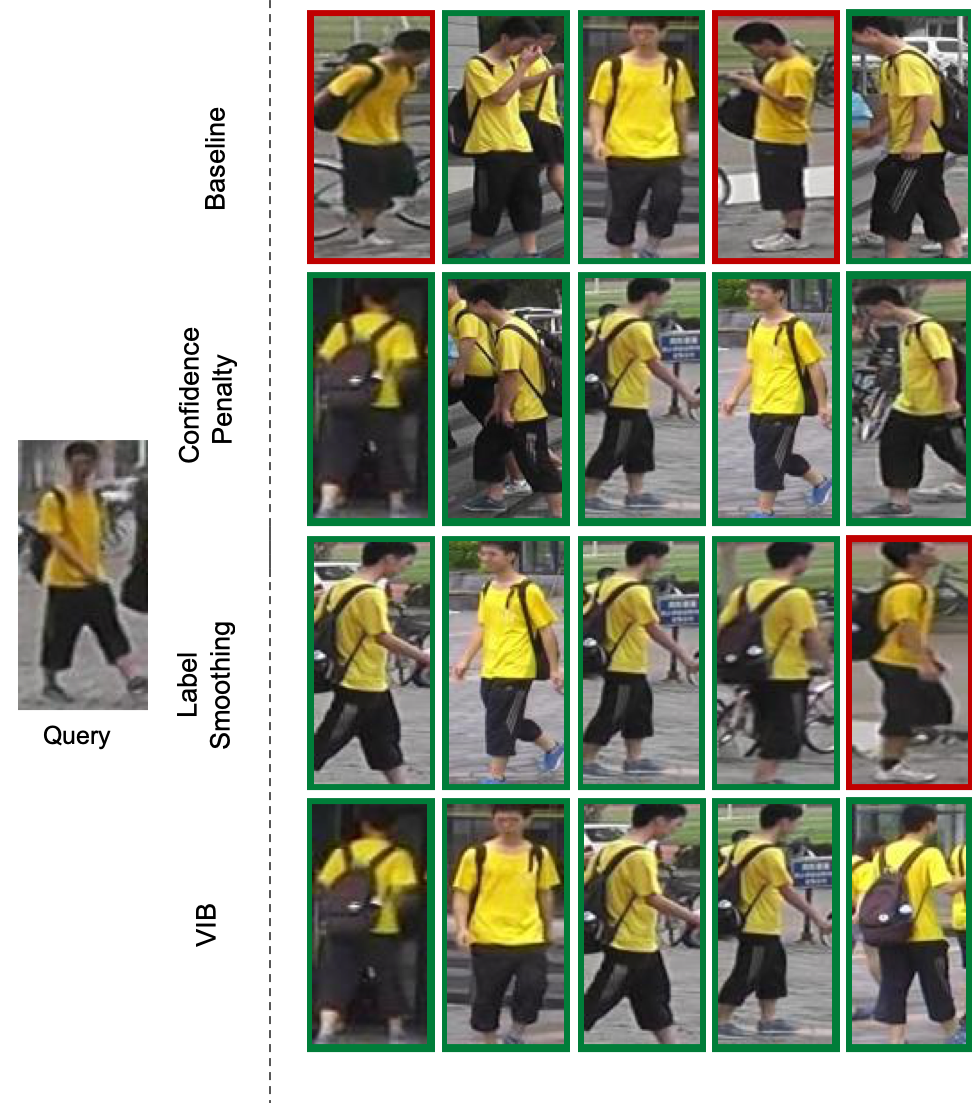}
    \caption{Qualitative comparison between the different methods on unseen Market1501 test samples. The gallery images are ranked according to L2 distance (top-5, left to right). Red frame indicates wrong ID while green frame indicates correct ID compared to the query. Best viewed in color.}
    % \label{fig:my_label}
\end{figure}

\begin{figure}
    \centering
    \includegraphics[width=1.05\linewidth]{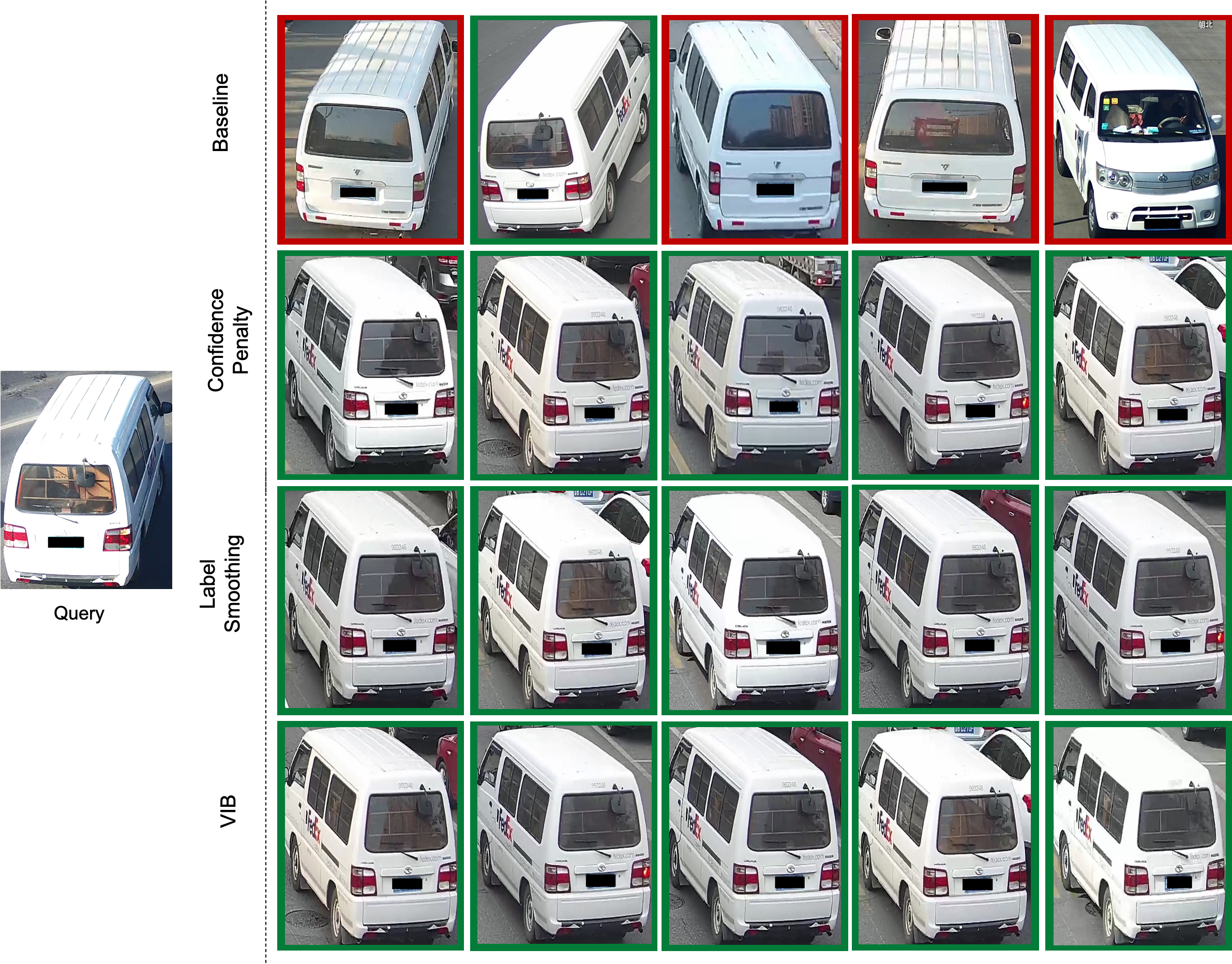}
    \caption{Qualitative comparison between the different methods on unseen VERI-Wild test samples. The gallery images are ranked according to L2 distance (top-5, left to right). Red frame indicates wrong ID while green frame indicates correct ID compared to the query. Best viewed in color.}
    % \label{fig:my_label}
\end{figure}

\begin{figure}
    \centering
    \includegraphics[width=1.05\linewidth]{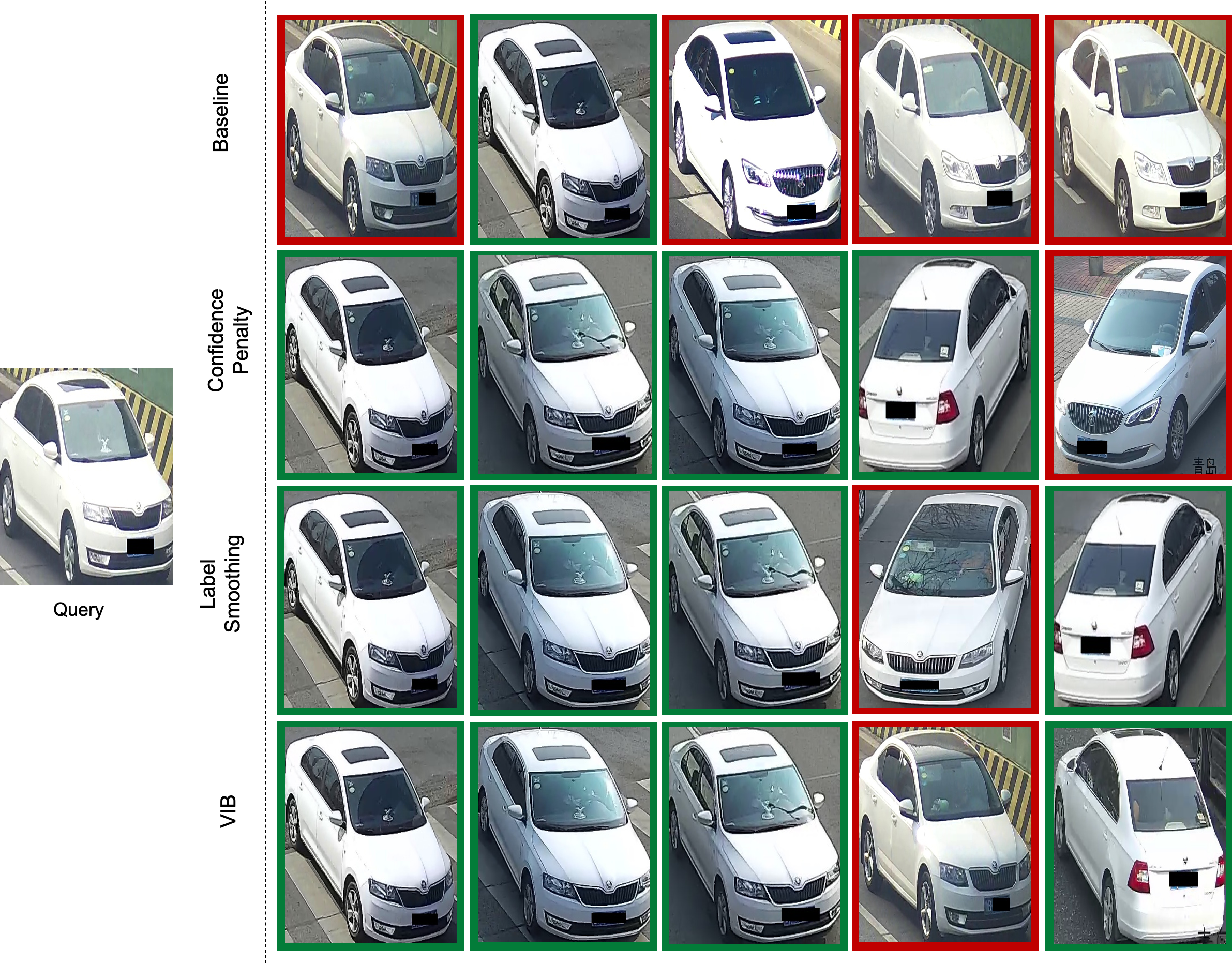}
    \caption{Qualitative comparison between the different methods on unseen VERI-Wild test samples. The gallery images are ranked according to L2 distance (top-5, left to right). Red frame indicates wrong ID while green frame indicates correct ID compared to the query. Best viewed in color.}
    % \label{fig:my_label}
\end{figure}
\end{document}